\documentclass{article}

\usepackage{microtype}
\usepackage{graphicx}
\usepackage{subfigure}
\usepackage{booktabs} 

\usepackage{hyperref}
\usepackage[noend]{algpseudocode}

\usepackage[accepted]{icml2024}

\usepackage{amsmath}
\usepackage{amssymb}
\usepackage{mathtools}
\usepackage{amsthm}

\usepackage[capitalize,noabbrev]{cleveref}

\theoremstyle{plain}
\newtheorem{theorem}{Theorem}[section]
\newtheorem{proposition}[theorem]{Proposition}
\newtheorem{lemma}[theorem]{Lemma}
\newtheorem{corollary}[theorem]{Corollary}
\theoremstyle{definition}

\theoremstyle{remark}
\newtheorem{remark}[theorem]{Remark}

\usepackage[textsize=tiny]{todonotes}
\usepackage{censor}
\usepackage{xcolor}
\usepackage{natbib}

\renewcommand{\l}{\left}
\renewcommand{\r}{\right}

\newcommand{\dd}{\text{d}}

\renewcommand{\algorithmiccomment}[1]{\hfill \normalsize $\triangleright$ ~#1\normalsize}

\icmltitlerunning{Liouville Flow Importance Sampler}

\begin{document}

\twocolumn[
\icmltitle{Liouville Flow Importance Sampler}



\icmlsetsymbol{equal}{*}

\begin{icmlauthorlist}
\icmlauthor{Yifeng Tian}{equal,yyy}
\icmlauthor{Nishant Panda}{yyy}
\icmlauthor{Yen Ting Lin}{equal,yyy}
\end{icmlauthorlist}

\icmlaffiliation{yyy}{Information Sciences Group (CCS-3), Computational and Statistical Sciences Division, Los Alamos National Laboratory, Los Alamos, NM 87545, USA}

\icmlcorrespondingauthor{Yifeng Tian}{yifengtian@lanl.gov}
\icmlcorrespondingauthor{Yen Ting Lin}{yentingl@lanl.gov}

\icmlkeywords{Probabilistic Machine Learning, Sampling Algorithms, Flow Models, ICML}

\vskip 0.3in
]



\printAffiliationsAndNotice{\icmlEqualContribution} 

\begin{abstract}

We present the Liouville Flow Importance Sampler (LFIS), an innovative flow-based model for generating samples from unnormalized density functions. LFIS learns a time-dependent velocity field that deterministically transports samples from a simple initial distribution to a complex target distribution, guided by a prescribed path of annealed distributions. The training of LFIS utilizes a unique method that enforces the structure of a derived partial differential equation to neural networks modeling velocity fields. By considering the neural velocity field as an importance sampler, sample weights can be computed through accumulating errors along the sample trajectories driven by neural velocity fields, ensuring unbiased and consistent estimation of statistical quantities. We demonstrate the effectiveness of LFIS through its application to a range of benchmark problems, on many of which LFIS achieved state-of-the-art performance.
\end{abstract}

\section{Background}

We are interested in sampling a hard-to-sample distribution $\nu$ in a continuous state space\footnote{We can also consider a subspace $ x\in \mathcal{E} \subseteq \mathbb{R}^D$.}  $x \in \mathbb{R}^D$, given its unnormalized probability density function $\tilde{\nu}(x)$ (assuming its existence, i.e., $\nu$ is dominated by the Lebesgue measure) up to a normalization constant $\mathcal{Z}$. We are also interested in estimating $\log \mathcal{Z}$, which is the log-marginal likelihood of a model and is a useful quantity for Bayesian model selection \citep{BayesFactors2,RJMCMC,llorenteMarginalLikelihoodComputation2023}. This problem arises in vast scientific domains, including statistical physics \cite{faulknerSamplingAlgorithmsStatistical2023}, molecular dynamics \citep{heninEnhancedSamplingMethods2022}, and diverse Bayesian inference and model selection problems, for example in computational and systems biology \citep{HINES20152103,wilkinsonBayesianMethodsBioinformatics2007}, astronomy \citep{sharmaMarkovChainMonte2017}, and political sciences \citep{jackmanEstimationInferenceBayesian2000,mccartanSequentialMonteCarlo2023}.

\subsection{Related Work}

Numerous Monte Carlo (MC) techniques have been devised to address this challenge over the past 70 years. Some notable examples of these MC methods include the landmark Metropolis Markov chain MC (MCMC, \citep{metropolisEquationStateCalculations1953}), Hybrid or Hamiltonian MC (HMC, \citet{duaneHybridMonteCarlo1987,nealMCMCUsingHamiltonian2012}), and the Zig-Zag Sampler \citep{bierkensZigZagProcessSuperefficient2019}. These approaches are often limited to low-dimensional problems due to slow convergence rate, and problems with a single mode as the samplers are prone to be trapped at a local mode. Annealed Importance Sampler (AIS, \citet{nealAnnealedImportanceSampling2001a}) and Sequential Monte Carlo (SMC, \cite{SMC}) were developed for sampling challenging multi-modal distributions and are considered the state-of-the-art MC methods for such problems. Variational Inference (VI, \citet{wainwrightGraphicalModelsExponential2008}) is an alternative approach, which turns the sampling problem into an optimization one by fitting an easy-to-sample parametric distribution to $\nu$, often by minimizing the reverse (or exclusive) Kullback--Leibler (KL) divergence. The choice of the reverse KL is commonly thought to be due to computational convenience but can be validated by an information theoretical argument of optimal information processing of Bayes' theorem \citep{zellnerOptimalInformationProcessing1988}. For VI, it is popular to adopt mean-field approximation \citep{wainwrightGraphicalModelsExponential2008} or a Normalizing Flow (VI-NF, \citet{VI-NF}) as the modeling parametric distribution. More recently, a plethora of neural network (NN) based algorithms, for example, neural network gradient HMC \cite{liNeuralNetworkGradient2019}, Annealed Flow Transport Monte Carlo (AFTMC, \citet{AFT}), Path-Integral Sampler (PIS, \citet{zhang2021path}), adaptive MC with NF \citet{gabrieAdaptiveMonteCarlo2022}, Continual Repeated AFTMC (CR-AFTMC, \citet{matthewsContinualRepeatedAnnealed2022}), and Denoising Diffusion Sampler (DDS, \cite{DDS}). For estimating the marginalized likelihood, simple MCMC methods require augmented algorithms (e.g., \citet{RJMCMC}) or additional processing of the derived samples (e.g., \citet{robertComputationalMethodsBayesian2009,pajorEstimatingMarginalLikelihood2017}). SIS and SMC can estimate the marginalized likelihood and are considered the gold-standard MC estimates. Among the NN-based algorithms, AFTMC, CR-AFTMC, PIS, and DDS are capable of estimating the marginalized likelihood. \citet{llorenteMarginalLikelihoodComputation2023} provide a more comprehensive review of methods estimating the marginalized likelihood. 

\begin{figure}[!t] 
    \centering
    \includegraphics[width=0.45\textwidth]{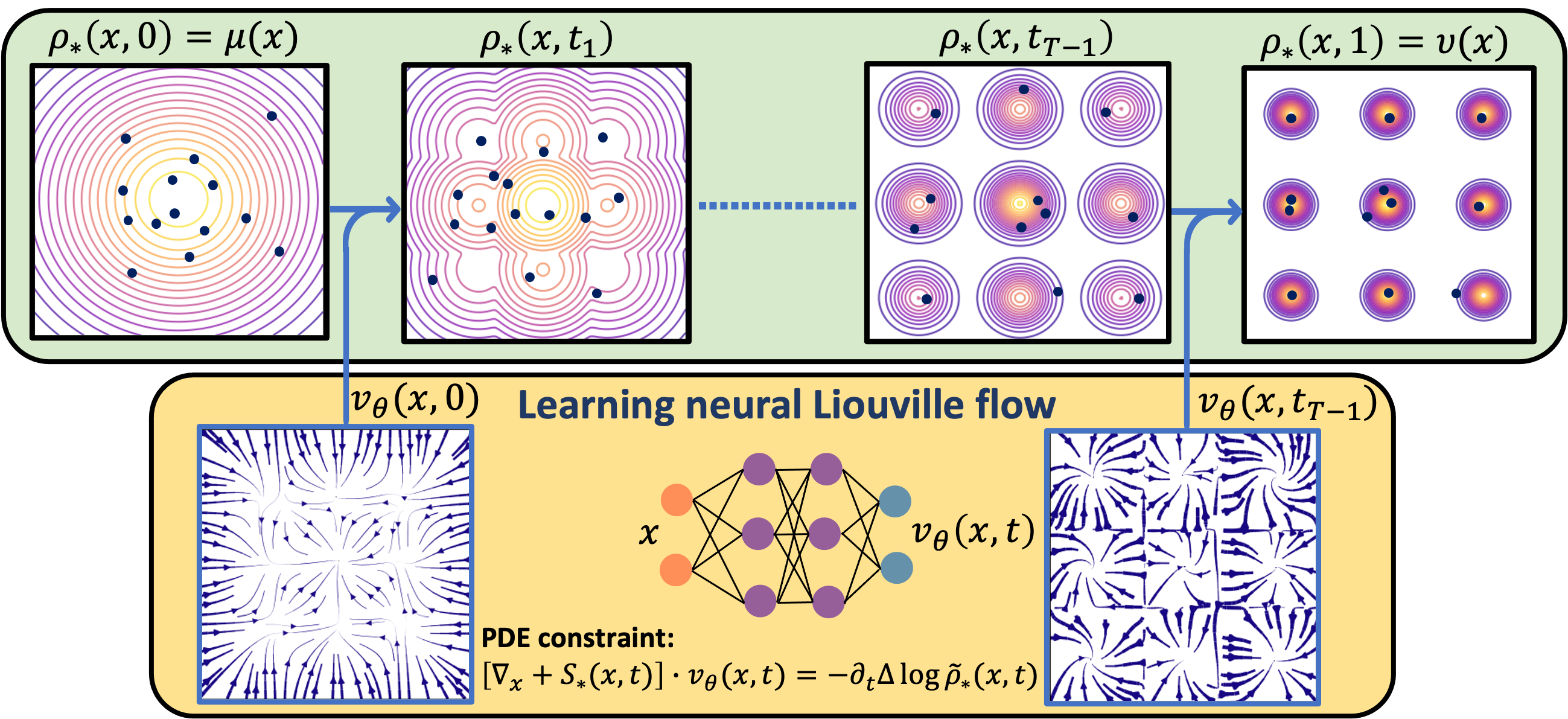}
    \caption{A schematic diagram demonstrating the workflow of the Liouville Flow Importance Sampler.}
    \label{fig:demo}
\end{figure}

\subsection{Summary of our contribution}

We propose \emph{Liouville Flow Importance Sampler}, a novel pure flow-based method with which we achieve sampling $\nu$ by (1) constructing a time-dependent target distribution\footnote{We use $\rho_\ast (t)$ to denote time-dependent distribution and $\rho_\ast (x,t)$ and $\tilde{\rho}_\ast(x,t)$ to denote its normalized and unnormalized density function respectively.} $\rho_\ast(t) $ connecting an easy-to-sample distribution $\mu$ and our target $\nu$, where $t\in \left[0,1\right]$ is a fictitious time, i.e., $\rho_\ast (0)=\mu$ and $\rho_\ast (1)=\nu$; (2) solving a time-dependent velocity field $v(x,t):\mathbb{R}^D \times \left[0,1\right] \rightarrow \mathbb{R}^D$ satisfying the following Generalized Liouville Equation (GLE, \citet{gerlich1973VerallgemeinerteLiouvilleGleichung}),
\begin{equation}
    \partial_t \rho_\ast (x,t) = -\nabla_x \cdot\left[ v(x,t) \rho_\ast(x,t)\right], \rho_\ast(x,0)=\mu(x) \label{eq:Liouville}
\end{equation}
(3)  drawing $N$ i.i.d.~samples $x_0^{(i)}$, $i=1\ldots N$ from $\mu$, then using the solved velocity field to evolve the samples from $t=0\rightarrow 1$ by the ordinary differential equation
\begin{equation}
    \dot{x}^{(i)}(t) = v\left(x^{(i)}(t), t\right),\ x^{(i)}(0)=x_0^{(i)}. \label{eq:ODE}
\end{equation}
The evolved samples at time $t=1$ are samples of $\nu$ because GLE ensures that the probability density of independently evolving samples is the density function in Eq.~\eqref{eq:Liouville}, i.e., $x^{(i)}(t) \sim \rho_\ast\left(t\right)$, and because $\rho_\ast(1)=\nu$. The schematic Fig.~\ref{fig:demo} illustrates the workflow of our proposal.

We summarize our novel contributions in this study, leaving a detailed discussion of the differences between our proposition and existing methods in the Results and Discussion section \ref{sec:discussion}:
 
$\bullet$ We introduce a pure flow-based model that deterministically transports samples from an initial distribution $\mu$ to a target distribution $\nu$, relying exclusively on the flow mechanism and consistent with a prescribed path of annealed distributions, $\tilde{\rho}_\ast(t)$. This innovation streamlines the modeling process, eliminating the complex tuning of meta and algorithmic parameters that is typical in hybrid models leveraging both flow and Monte Carlo methods.
    
$\bullet$ We propose an original approach to solving a key equation (Eq.~\eqref{eq:consistent} below), which $v(x,t)$ must satisfy to ensure the samples move consistently with the specified time-dependent and unnormalized density function $\tilde{\rho}_\ast(x,t)$. We convert the problem to an equation-based machine learning task where samples drawn from $\mu$ are used to train an NN that models the velocity field at a specific time. Trained NNs are used to generate new samples for training the velocity field at a later time. This recursive training of a series of NNs facilitates end-to-end sample transportation from $\mu$ to $\nu$. 

$\bullet$ We demonstrate a novel derivation that, although finite NNs do not perfectly learn the solution of Eq.~\eqref{eq:consistent}, the accumulated error induced  along the sample trajectory can be used as sample weights, allowing for unbiased and consistent estimation of statistical quantities. 

$\bullet$ The proposed method, while having a simpler setup compared to existing methods, achieves state-of-the-art performance across a variety of standard benchmarks.

\section{Liouville Flow Importance Sampler}
\subsection{Governing equation of the velocity field}

Our proposition hinges on the key research question: Given an unnormalized time-dependent density function $\tilde{\rho}_\ast(x,t) = \mathcal{Z}(t) \rho_\ast(x,t)$, up to an unknown normalization constant $\mathcal{Z}(t):= \int \tilde{\rho}_\ast\left(x,t\right) \dd x$, how do we solve a time-dependent velocity field $v(x,t)$? 

We first ask the question: what is the equation a velocity field $v(x,t)$ must satisfy in terms of only the unnormalized density function $\tilde{\rho}_\ast(x,t)$ (instead of the inaccessible full density function $\rho(x,t)$ in Eq.~\eqref{eq:Liouville})? On the one hand, using GLE~\eqref{eq:Liouville} and the chain rule, it is straightforward to establish
\begin{equation}
    \partial_t \log \rho_\ast(x,t) = -\nabla_x \cdot v\left(x,t\right)  - v \left(x,t\right) \cdot S_\ast \left(x,t\right), \label{eq:dlogrhoastdt1}
\end{equation}
where $S_\ast\left(x,t\right):=\nabla_x \log \rho_\ast \left( x, t \right)\equiv \nabla_x \log \tilde{\rho}_\ast \left( x, t \right)$ is the score function of the target density at time $t$. Here, we assume that the derivative of the log-density exists and can be evaluated.
On the other hand, applying $\partial_t$ to $\log\rho_\ast\left(x,t\right)= \log\tilde{\rho}_\ast \left(x,t\right) - \log \mathcal{Z}\left(t\right) $ leads to 
\begin{equation}
    \partial_t \log \rho_\ast(x,t) = \partial_t \log \tilde{\rho}_\ast(x,t) - \left\langle \partial_t \log \tilde{\rho}_\ast(\cdot,t) \right \rangle_\ast, \label{eq:dlogrhoastdt2}
\end{equation}
where we used a common notation in statistical physics, $ \left\langle \partial_t \log \tilde{\rho}_\ast(\cdot,t) \right \rangle_\ast :=\mathbb{E}_{x'\sim \rho_\ast(\cdot, t)} \left[\partial_t \log \tilde{\rho}_\ast(x',t) \right] $. This term emerged from the time-derivative of the unknown normalization constant $\mathcal{Z}(t)$. Equating Eqs.~\eqref{eq:dlogrhoastdt1} and \eqref{eq:dlogrhoastdt2}, we obtained the following equation:
\begin{subequations}\label{eq:consistent}
\begin{align}
    {}& \left[\nabla_x + S_\ast\left(x,t\right) \right]\cdot v \left(x,t\right) = - \partial_t \delta \log \tilde{\rho}_\ast(x,t), \\
    {}& \delta \log \tilde{\rho}_\ast(x,t) := \log \tilde{\rho}_\ast(x,t)-\left\langle \log \tilde{\rho}_\ast(\cdot,t) \right \rangle_\ast. \label{eq:source}
\end{align}
\end{subequations}
The below theorem, whose proof is given in Appendix \ref{app:existenceProof}, states the existence of a solution to Eq.~\eqref{eq:consistent}.
\begin{theorem}\label{thm:existence_vel_field}
Suppose $\partial_t \delta \log \tilde{\rho}_\ast \left(x,t\right)$ is square-integrable and the divergence of target score function is absolutely bounded i.e., $\exists M < \infty$ s.t.~$\lvert\nabla_x\cdot S_{\ast}(x,t)\rvert < M $ almost everywhere for all $t$.
Then, there exists a solution of Eq.~\eqref{eq:consistent}, but the solution is not unique in $D\ge 2$ dimensions.
\end{theorem}
We remark that the non-uniqueness of the solution does not pose an issue for our objectives. Any solution of $v(x,t)$ satisfying Eq.~\eqref{eq:consistent}, once obtained, has an equivalent effect of transporting the samples.

\subsection{Learning velocity field for sampling} \label{sec:idealLearning}
It is beyond the authors' knowledge how to solve Eq.~\eqref{eq:consistent} accurately in an arbitrary dimension $D$. As such, instead of solving the equation, we propose to \emph{learn} a velocity field that approximately satisfies Eq.~\eqref{eq:consistent} by the following procedure. Suppose we could generate samples $x^{(i)}$ at time $t$ from $\rho_\ast\left(t\right)$. Both $S_\ast(x^{(i)},t)$ and $\partial_t \log \tilde{\rho}_\ast(x^{(i)},t)$ can be evaluated, and $\left\langle  \partial_t \log \tilde{\rho}_\ast(\cdot,t) \right \rangle_\ast$ can be estimated by Monte Carlo, i.e., $(1/N) \sum_{i=1}^N \partial_t \log\tilde{\rho}_\ast(x^{(i)},t)$. Next, we use an NN to model the velocity field at this specific time $t$, i.e. $v_\theta (x,t) = \text{NN}(x;\theta,t)$ with NN parameters $\theta$. Using automatic differentiation, the divergence term, $\nabla_x \cdot v_\theta(x^{(i)},t)$ can be evaluated for each sample. Replacing the solution $v(x,t)$ in Eq.~\eqref{eq:consistent} with the neural velocity field $v_\theta(x^{(i)},t)$, our objective is to minimize the discrepancy of the LHS and the RHS in Eq.~\eqref{eq:consistent} over the samples at time $t$, 
\begin{subequations} \label{eq:training}
\begin{align}
    \theta_\ast ={}& \text{argmin}_\theta  \sum_{i=1}^N \varepsilon^2\left(x^{(i)};\theta\right),\\
    \varepsilon\left(x;\theta\right):={}& \left[\nabla_x + S_\ast\left(x\right) \right]\cdot v_\theta \left(x\right) + \partial_t \delta \log \tilde{\rho}_{\ast}\left(x\right), \label{eq:staticError}
\end{align}
\end{subequations}
where we have suppressed the $t$-dependence of $S_\ast$, $v_\theta$, and $\tilde{\rho}$ for brevity. After the learning, we use the learned velocity field to evolve the sample to a later time $t+\Delta t$ by an explicit Euler scheme, $x^{(i)}\leftarrow x^{(i)} + v_{\theta_{\ast}}\left(x^{(i)}, t\right)\times \Delta t$, $\Delta t \ll 1$. In the ideal scenario that the NN learns the solution of Eq.~\eqref{eq:consistent} and the error induced by the time discretization is negligible, the GLE \eqref{eq:Liouville} ensures that the transported samples $\sim \rho_\ast \left(t+\dd t\right)$. In turn, the samples can be used to learn the velocity field at $t+\Delta t$. Repeating the process, the samples can be recursively evolved from $t=0\rightarrow 1$.   

Our proposition of learning the velocity field using equation Eq.~\eqref{eq:consistent} is similar to the Physics Informed Machine Learning \citep{RAISSI2019686,karniadakisPhysicsinformedMachineLearning2021}. Moreover, our proposition can be considered as a self-learning framework because samples are not labeled, and because we can always generate more samples from $\mu$ and evolve them by previously trained $v(x,s)$ to $t>s$, for training $v(x,t)$.  We note that Eq.\eqref{eq:consistent} has been explicitly presented in \citet{Jarzynsky08}, \citet{hengGibbsFlowApproximate2021}, and \citet{AFT}; a more detailed discussion is provided in Sec.~\ref{sec:discussion}. 

We remark that our methodology significantly differs from the Flow Matching approach \cite{lipman2023flow,tong2024improving}. Flow Matching trains a continuous-time normalizing flow by matching it to a \textit{known} velocity field. This technique is specifically designed for training Generative Models using samples of a data distribution. In contrast, our task involves learning the normalizing flow using a prescribed path of annealed distributions, $\tilde{\rho}_\ast (x,t)$. In this scenario, we do not have access to the samples of the target distribution nor a target velocity field for matching: the problem would have been solved if we already had the samples of the target distribution, and we could have simply driven the samples by a known target velocity field if it is already known. Instead, our knowledge is limited to the fact that the velocity field at each of the samples must satisfy Eq.~\eqref{eq:consistent}. Our goal is to solve for the unknown $v(x,t)$ in Eq.~\eqref{eq:consistent}, rather than attempting to match a target flow.

\subsection{Neural network as an importance sampler} \label{sec:IS}

Our proposition described in Sec.~\ref{sec:idealLearning} works \emph{only if} the NN is expressive enough to accurately reproduce the true solution satisfying Eq.~\eqref{eq:consistent}. Without the assumption, transported samples would not be representative samples at $t+\dd t$ for training $v_\theta$. 

More specifically, let us consider a sub-optimal velocity field $v_{\theta_\ast}\left(x,t\right)$ which does not satisfy Eq.~\eqref{eq:consistent}, $\forall x\in \mathbb{R}^D$. This is very likely because the training of an NN is not perfect, or because we may not have enough computational resources to train an expressive enough network to fully solve Eq.~\eqref{eq:consistent}. Using the trained NNs to transform the initial distribution $\mu$, the induced distribution $\rho_\theta(t)$ still satisfies a GLE $\partial_t \rho_\theta(x,t) = - \nabla_x \left[v(x,t) \rho_\theta\left(x,t\right)\right]$ but $\rho_\theta(t) $ is no longer be the same as $\rho_\ast(t)$. In this case, the RHS of Eq.~\eqref{eq:dlogrhoastdt1} becomes $-\nabla_x \cdot v_\theta \left(x,t\right)  -  S_\theta \left(x,t\right) \cdot v_\theta \left(x,t\right)$, where $S_\theta:= \nabla_x \log \rho_\theta\left(x,t\right)$, and consequently we could not establish Eq.~\eqref{eq:consistent}. In addition, we will no longer have samples to estimate unbiasedly the term $\left\langle \partial_t \log \tilde{\rho}_\ast(\cdot,t) \right \rangle_\ast$ in Eq.~\eqref{eq:consistent}. Our proposition outlined in Sec.~\ref{sec:idealLearning} appears to lose its theoretical grounding.

Fortunately, a more careful derivation presented in Appendix \ref{app:derivations} revealed that we can use the trained, albeit imperfect, neural networks to transport the samplers as an importance sampler. We provide key equations here, leaving detailed derivation and proof in Appendices \ref{app:derivations} and \ref{app:logZ}. 

We first write the generic solution of Eq.~\eqref{eq:ODE}:
\begin{equation}
    x^{(i)}(t) = x_0^{(i)} + \int_{0}^t v_\theta\left(x(s;x_0^{(i)}), s\right) \, \dd s,\label{eq:sample_traj_nn}
\end{equation}
where $v_\theta$ is an NN-modeled velocity field and $x_0^{(i)} \sim \mu$ is the initial condition. Below, when the context is clear, we will suppress the sample superscript $(i)$ and write $x(t)$ and $x_0$ for brevity. We stress that the sample trajectory implicitly depends on the initial condition $x_0$, which is the only randomness; once $x_0$ is drawn, $x(t)$ is a deterministic trajectory. 
Then, the probability density of the modeling distribution along this trajectory can be computed (see Appendix \ref{app:derivations}):
\begin{equation}
    \log \rho_\theta(x(t), t) =\log \mu(x_0) - \int_0^t \nabla \cdot v_\theta\left(x(s),s\right)\, \dd s.  \label{eq:logrhotheta-main} 
\end{equation}
Treating $\rho_\theta$ as an importance sampler, interestingly, the unnormalized log-weights of the samplers can be shown to be the error accumulated along each of the sample trajectories:
\begin{equation}
    \log \tilde{w}\left(t;x_0\right) =  \int_{0}^t \varepsilon\left(x\left(s;x_0\right)\right) \,  \dd s, \label{eq:log-weights}
\end{equation}
where $\varepsilon$ is defined in Eq.~\eqref{eq:staticError}. With $N$ initial samples $x_{0}^{(i)} \sim \mu$, $i=1\ldots N$, we can obtained the normalized weights \citep{liuMonteCarloStrategies2008,tokdar2010importance} 
\begin{equation}
{w}_i(t) := { \tilde{w}\left(t, x_0^{(i)}\right)}/{ \sum_{j=1}^{N} \tilde{w}\left(t, x_0^{(j)}\right)}. \label{eq:weights-main}
\end{equation}
which can be used to estimate unbiasedly asymptotically (i.e.~unbiased as $N\rightarrow \infty$)
\begin{equation}
    \left\langle \partial_t \log \tilde{\rho}_\ast(\cdot,t) \right \rangle_\ast \approx \sum_{i=1}^N  {w}_i (t) \partial_t \log \tilde{\rho}_\ast\left(x^{(i)}\left(t\right) ,t\right),
\end{equation}
which can be used for learning $v(x,t)$ without an accurate solution of Eq.~\eqref{eq:consistent}. Furthermore, a derivation akin to \citet{gelmanSimulatingNormalizingConstants1998} revealed that we can unbiasedly asymptotically estimate the logarithm of 
 the marginalized likelihood $\log \mathcal{Z}(t)$,
\begin{equation}
    \widehat{\log \mathcal{Z}(t)} \approx \int_0^t \sum_{i=1}^N  {w}_i \left(s\right) \partial_s \log \tilde{\rho}_\ast\left(x^{(i)}\left(s\right) ,s\right) \dd s, \label{eq:logZestimator}
\end{equation}
which can be computed on-the-fly as samples are transported from $t=0\rightarrow 1$. Theorem \ref{thm:logZ}, whose proof is given in Appendix \ref{app:logZ}, articulates the mathematical statements about the asymptotically unbiased estimator. 
\begin{theorem}\label{thm:logZ}
   For any $t\in [0,1]$, suppose the $\partial_t \log \tilde{\rho}_\ast(x,t)$ is an integrable function with respect to the target distribution $\rho_\ast(t)$. Let $\lbrace x_0^{(i)}\rbrace_{i=1}^N$ be a finite set of initial samples drawn from the initial distribution $\mu$ and consider the following finite sum for each time $s \in (0,t)$ given by
    \begin{equation} 
        W_N(s) = \sum\limits_{i=1}^N w_i\left(s\right) \partial_t \log \tilde{\rho}_\ast\left(x^{(i)}(s),s\right)
    \end{equation}
    where $w_i\left(s\right)$ is the dynamic sample weight along the trajectory given by the evolution of $x_0^{(i)}$ as defined in Eqs.~\eqref{eq:log-weights} and \eqref{eq:weights-main}. Suppose the following conditions hold: (1) the neural velocity field induced distribution $\rho_\theta(t)$ dominates the target distribution $\rho_\ast(t)$ and, (2) both the dynamic weights $\tilde{w}\left(t;x_0\right)$ and $\partial_t \log\tilde\rho$ are absolutely bounded, i.e there exists constants $M_1, M_2$ such that $\lvert\tilde{w}\left(t;x_0\right)\rvert < M_1$ and $\lvert \partial_{t} \log \tilde{\rho}_\ast \left(x,t\right)\rvert < M_2$. Then, the marginalized likelihood can be estimated unbiasedly asymptotically (i.e.~unbiased as $N\rightarrow \infty$) and persistently 
    \begin{align*}
      \int_{0}^tW_N(s)\, \dd s \overset{\text{a.s.}}{\longrightarrow} \log \mathcal{Z}(t) \text{ as } N\to \infty.
    \end{align*}
\end{theorem}

{While the above theorem provides the unbiased estimation of $\log\mathcal{Z}$ directly, i.e., $\widehat{\log \mathcal{Z}}$, many published methods focus on unbiasedly estimating $\mathcal{Z}$ directly, i.e., $\hat{\mathcal{Z}}$, but report $\log \hat{\mathcal{Z}}$. To make a fair comparison, we established a similar estimator for LFIS in the following theorem
\begin{theorem}\label{thm:Z}
    For any $t\in [0,1]$, suppose $\partial_t \log \tilde{\rho}_\ast(x,t)$ is an integrable function with respect to the target distribution $\rho_\ast(t)$. Let
    $\epsilon\left(x;\theta\right)$ be a modified error function that is analogous to Eq.~\eqref{eq:staticError} as 
\begin{equation}
\epsilon\left(x;\theta\right):= \left[\nabla_x + S_\ast\left(x\right) \right]\cdot v_\theta \left(x\right) + \partial_t \log \tilde{\rho}_\ast\left(x\right), \label{eq:modifiedError}
\end{equation}
and the modified unnormalized importance weights that are analogous to Eq.~\eqref{eq:log-weights}:
\begin{equation}
    \log \varpi\left(t;x_0\right) =  \int_{0}^t \epsilon\left(x\left(s;x_0\right)\right) \dd s. \label{eq:modifiedLogWeights}
\end{equation}
If the same conditions as in Theorem~\ref{thm:logZ} hold, the marginalized likelihood $\mathcal{Z}$ can be unbiasedly estimated:
\begin{equation}
  \hat{\mathcal{Z}}\left(t\right) = \frac{1}{N}\sum_{i=1}^N e^{\log\varpi\left(t;x_0\right)}.  \label{eq:Zestimate}
\end{equation}
\end{theorem}
The proof of the above estimate is given in Appendix \ref{app:logZ}. In the rest of the main text, we focus on $\log \hat{\mathcal{Z}} $ to ensure a fair comparison between different methods and report $\widehat{\log \mathcal{Z}}$ in the Appendix (Table \ref{tab:LFIS-all}).

\begin{algorithm*}[!t]
\caption{Liouville Flow Importance Sampler}\label{alg:1}
\begin{algorithmic}
\Procedure{GenerateSamples}{$k$, $n$, $\theta_\ast^{(0:k-1)}$, optional $\langle \partial_t \log \tilde{\rho}_\ast \rangle_\ast^{(0:k-1)}$}
    \State \Comment{Generate $n$ samples at a target time $k/T$ using previously trained $v_{\theta_\ast^{(j)}}$ and optionally provided $\langle \partial_t \log \tilde{\rho}_\ast \rangle_\ast^{(0:k-1)}$}
    \State $x_i \sim \mu$, $\delta_i \leftarrow 0$, $i=1\ldots n$     \Comment{Drawing samples from the initial distribution} 
    \For{$\ell=0\ldots k-1$}  \Comment{Evolve samples using previously trained fields $v_{\theta_\ast}$ and estimated $\langle \partial_t \log \tilde{\rho}_\ast \rangle_\ast$}
        \State {\bf If} $\langle \partial_t \log \tilde{\rho}_\ast \rangle_\ast^{(\ell)}$ is not provided {\bf then} $\langle \partial_t \log \tilde{\rho}_\ast \rangle_\ast^{(\ell)} \leftarrow \sum_{i=1}^n \exp(-\delta_i) \partial_t \log \tilde{\rho}_\ast\left( x_i , \ell/T \right) /\left(T \sum_{j=1}^n \exp(-\delta_j)\right)$  
        \State $\delta_i \leftarrow \delta_i + \left\{\left[ \nabla_x+ S_\ast \left(x_i,{\ell}/{T}\right)\right]\cdot v_{\theta_\ast^{(\ell)}} \left(x_i\right) + \partial_t \log \tilde{\rho}_\ast\left(x_i,{\ell}/{T}\right) - \langle \partial_t \log \tilde{\rho}_\ast \rangle_\ast^{(\ell)}\right\}/T$ \Comment{Error accumulation}
        \State $x_i \leftarrow x_i + v_{\theta_\ast^{(\ell)}}\left(x_i\right)/T$ \Comment{Transporting samples}
    \EndFor
    \State $w_i \leftarrow \exp(-\delta_i) / \sum_{j=1}^n \exp(-\delta_j)$ \Comment{Weight computation}
    \State \Return $\left\{x_i, w_i\right\}_{i=1}^N$
\EndProcedure
\Procedure{Learning}{~} \Comment{Learning the flow}
    \For{$k=0 \ldots T-1$}
        \State $\left\{x_i, w_i\right\}_{i=1}^N=\text{GenerateSamples}\left(k,\, N,\, \theta_\ast^{(0:k-1)},\, \langle \partial_t \log \tilde{\rho}_\ast \rangle_\ast^{(0:k-1)}\right)$  \Comment{Samples for estimating $\langle \partial_t \log \tilde{\rho}_\ast \rangle_\ast^{(k)}$}
        \State $\langle \partial_t \log \tilde{\rho}_\ast \rangle_\ast^{(k)} \leftarrow \sum_{i=1}^N w_i\partial_t \log \tilde{\rho}_\ast\left( x_i ,k/T \right) $ \Comment{Store the estimated $\langle \partial_t \log \tilde{\rho}_\ast \rangle_\ast^{(k)}$}
        \While{Training criteria are not met} \Comment{Trianing the NN}
            \State $\left\{x_i, w_i\right\}_{i=1}^B=\text{GenerateSamples}\left(k,\, B,\, \theta_\ast^{(0:k-1)},\, \langle \partial_t \log \tilde{\rho}_\ast \rangle_\ast^{(0:k-1)}\right)$  \Comment{Batch ($B$) samples for training}
            \State $\varepsilon_i(\theta)  \leftarrow \left[ \nabla_x+ S_\ast \left(x_i,k/T\right)\right]\cdot v_{\theta} \left(x_i\right) + \partial_t \log \tilde{\rho}_\ast(x_i,k/T) - \langle \partial_t \log \tilde{\rho}_\ast \rangle_\ast^{(k)}$ \Comment{Sample-by-sample error}
            \State $\theta \leftarrow \text{GradientDecentStep}\left(\theta, \nabla_\theta \left(\sum_{i=1}^{B} \varepsilon_i^2 (\theta)/B\right)\right)$
        \EndWhile
        \State $\theta_\ast^{(k)} \leftarrow \theta$ \Comment{Store trained NN parameters}
    \EndFor
\EndProcedure
\Procedure{Sampling}{} \Comment{Generate $S$ samples of $\nu$ after learning the flow}
    \State \Return{$\text{GenerateSamples}\left(T,\, S,\, \theta_\ast^{(0:T-1)}\right)$}
\EndProcedure
\end{algorithmic}
\end{algorithm*}

\subsection{Choice of $\tilde{\rho}_\ast(x,t)$ and the schedule function $\tau(t)$} \label{sec:rho_ast(t)}

Our proposition requires an unnormalized time-dependent target density function $\tilde{\rho}_\ast(x,t)$. In this manuscript, we primarily consider two types of applications: (1) sampling a distribution $\nu$, given its unnormalized density function $\tilde{\nu}(x)$ and (2) Bayesian posterior sampling, given a prior density function $\pi(x)$ and a likelihood function $L(x)$. Motivated by AIS \citep{nealSamplingMultimodalDistributions1996,nealAnnealedImportanceSampling2001a} and SMC \citep{SMC}, we consider $\tilde{\rho}_\ast(x,t) := \mu^{1-\tau\left(t\right)}(x)\,  \tilde{\nu}^{\tau\left(t\right)} (x)$ for application (1) and $\tilde{\rho}_\ast(x,t) := L^{\tau\left(t\right)} (x) \pi(x)$ for application (2), where $\tau(t)$ is a monotonic function transforming time $t$, satisfying $\tau(0)=0$ and $\tau(1)=1$. We term $\tau(t)$ as the \emph{schedule function}. The corresponding key quantities are
\begin{subequations}
\begin{align}
    {}& \partial_t \log \tilde{\rho}_\ast(x,t) = \left[ \log \tilde{\nu}(x) - \log \mu (x)\right] \frac{\dd \tau\left(t\right)}{\dd t},  \nonumber\\
    {}& S_\ast(x,t) = \left(1-\tau \left(t\right)\right) \nabla_x \log \mu(x) + \tau \left(t\right) \nabla_x \log \tilde{\nu}(x)\nonumber
\end{align}
\end{subequations}
for type-1 application, and 
\begin{subequations}
\begin{align}
    {}& \partial_t \log \tilde{\rho}_\ast(x,t) = \log L(x) \frac{\dd \tau\left(t\right)}{\dd t}, \nonumber\\
    {}& S_\ast(x,t) =  \tau\left(t\right) \nabla_x \log L(x) + \log \pi(x)\nonumber
\end{align}
\end{subequations}
for application type-2. We remark that the general learning and sampling procedures of LFIS do not depend on the particular choice of $\tilde{\rho}_\ast(t)$. 

We term our proposition in Sec.~\ref{sec:idealLearning} with the importance sampling in Sec.~\ref{sec:IS} and the choices of $\tilde{\rho}_\ast(x,t)$ in Sec.~\ref{sec:rho_ast(t)} as the \emph{Liouville Flow Importance Sampler (LFIS)}. 

\section{Numerical Experiments} \label{sec:numerical}

In this section, we present numerical experiments comparing the proposed LFIS with other state-of-the-art approaches using NNs, including AFTMC, PIS, and DDS. We left out VI-NF as its inferior performance has been established in \citet{AFT}. We also optimized an SMC to provide a reference of the performance of a state-of-the-art sampling algorithm without NNs; see Appendix \ref{app:SMC}.

We use a set of NNs to model the temporally discretized velocity field $v(x,t)$, at $t=0,1/T, 2/T,\ldots, 1$. Here, $T$ is the total number of time steps, which plays the same role as the number of tempering scales (or ``temperatures'') in AIS, AFT, or SMC. We investigated $T=32$, $64$, $128$, and $256$. Results with $T=256$ of all the methods are presented in the main manuscript, and others are presented in Appendix \ref{app:Tsteps} for completeness. We chose a cosine schedule function (see Sec.~\ref{sec:rho_ast(t)}) based on the results of a smaller-scale analysis presented in Appendix \ref{app:scheduling}. At each discrete time step, we used a separate feed-forward NN with a similar structure as in \citet{DDS} and \citet{zhang2021path} (two hidden layers, each of which has 64 nodes) to model the discrete-time velocity field. Except for the first ($t=0$) NN, which was initialized randomly, we instantiated the NN at $t=k/T$ using the weights of the trained NN at the previous time $t=(k-1)/T$ to amortize the training cost. We initialized the weights of the last layer of the NN to be zero, which was observed to expedite the training process empirically. The divergence of the flow field and the score function can either be computed theoretically or by using the \texttt{autograd} function in \texttt{PyTorch}. 

Algorithm \ref{alg:1} provides a more detailed description of the implementation of LFIS. Samples are drawn from either standard isotropic Gaussian distribution for type-1 applications, or from the prior distribution for type-2 applications (see Sec.~\ref{sec:rho_ast(t)}). The samples are transported by the previously learned velocity fields to a specific time step, which will then be used for learning the velocity field at the next time step. Our experiments suggested that the learning performs better with the following implementations. (1) A large number of samples ($N=5\times 10^4$) are used to estimate $\left\langle \partial_t \log \tilde{\rho}_\ast \left(t\right) \right \rangle_\ast$ at a specific time. For computational efficiency, we only draw a fixed batch at each target time to estimate this quantity once. (2) A small batch of samples ($B=300$-$10000$ depending on applications) are drawn for each gradient descent step.

After training, we performed 30 independent samplings (each with $S=2000$ samples, different from those used in training) using the learned velocity fields. Following the recent series of papers \citep{AFT,zhang2021path,DDS}, we focus on assessing sample qualities by the estimation of log normalization constant $\log\hat{\mathcal{Z}}$. For low dimensional ($D\le 10$) problems, we also compare the sliced Wasserstein distance $W_p$ \citep{bonneelSlicedRadonWasserstein2015} to ground-truth samples with $p=2$ as an additional metric. For all the methods with importance sampling, we used the weighted samples for computing these statistical quantities.

\subsection{Testing problems}

\textbf{Mode-separated Gaussian mixture (type-1)}: An illustrative model of $D=2$ Gaussian mixture distribution with separated modes. We consider a similar challenging distribution as in PIS \citep{zhang2021path}: a mixture of nine Gaussian distributions centered at the grid $\{-1,0,1\}^2$, and each Gaussian has variance 0.012. We present results with equally-weighted ($1/9$) modes in the main text, and unequally-weighted modes in Appendix \ref{app:MG2DWeights}.

\textbf{Funnel distribution (type-1)}: A challenging ten-dimensional distribution proposed by \citet{nealSliceSampling2003} is commonly used for testing samplers. The formulation of the funnel distribution is:
\begin{subequations}
    \label{eq:funnel}
    \begin{align}
    x_0 \sim{}& \mathcal{N}(\mu=0,\sigma^2 = 9),\\ x_{1:9}\vert x_0
 \sim{}& \mathcal{N}(\boldsymbol{\mu}=\mathbf{0},\boldsymbol{\Sigma}=e^{x_0} \mathbf{I})
    \end{align}
\end{subequations}

\textbf{Log Gaussian Cox Process (type-2)}: The Log-Gaussian Cox Process (LGCP) is a commonly used model for the analysis of spatial point pattern data and is designed for modeling the positions of Findland pine saplings \citep{LGCP}. The LGCP is a hierarchical combination of a Poisson process and a Gaussian Process prior, which can be naturally framed as a Bayesian problem. Here we use the variant of LGCP on a $40\times40$ grid, resulting a $D=1600$ sampling problem. The target posterior density is: 
\begin{equation}
\label{eq:LGCP}
\lambda(\mathbf{x}) \sim \mathcal{N}(\mathbf{x}; \mu, K)\prod_{i}\exp(x_iy_i - \alpha e^{x_i}).
\end{equation}
\begin{table}[!t]
\caption{{Results of type-1 problems. The best model (in bold font) in $\log \hat{\mathcal{Z}}$ is determined by its deviation from the theoretical value, $0$. The best model in Wasserstein-2 distance to the ground-truth samples is determined by the smallest value. The best model in effective sample size (EES) is determined by its deviation from the optimal value, $1$. }}
\vskip 0.15in
\centering
\begin{small}
\begin{tabular}{c c c c}
\toprule
& Model  & MG ($D=2$) & Funnel ($D=10$) \\ 
\midrule 
&SMC  & -1.28 $\pm$ 0.01 & -0.12  $\pm$ 0.06 \\
& {LFIS} & \bf{-0.0002 $\pm$ 0.004} & {\bf{-0.07 $\pm$ 0.003}} \\
$\log \hat{\mathcal{Z}}$& DDS  & -0.31 $\pm$ 0.43 &-0.31$\pm$ 0.12 \\
&PIS &  0.0035 $\pm$ 0.02& -1.14 $\pm$ 0.13  \\
&AFT & 2.43 $\pm$ 0.05 & -0.11 $\pm$ 0.68 \\
\midrule
&SMC & 0.066 $\pm$ 0.017 & 6.07 $\pm$ 1.52 \\
&LFIS & \bf{0.054 $\pm$ 0.014} & \bf{5.57 $\pm$ 1.52}\\
$W_2$ &DDS & 0.368 $\pm$ 0.23 & 6.54 $\pm$ 1.44\\
&PIS &0.061 $\pm$ 0.015 & 6.40 $\pm$ 1.49 \\
&AFT & 0.10 $\pm$ 0.042 & 6.20 $\pm$ 1.41 \\
\midrule
&{SMC} & {0.99 $\pm$ 0.003} & {0.99 $\pm$ 0.005} \\
&{LFIS} & {\bf{0.97 $\pm$ 0.001}} & {\bf{0.97 $\pm$ 0.063}}\\
ESS &{DDS} & {0.012 $\pm$ 0.009} & {0.163 $\pm$ 0.086}\\
&{PIS} &{0.59 $\pm$ 0.038} & {0.12 $\pm$ 0.066}\\
&{AFT} & {0.66 $\pm$ 0.185} & {0.76 $\pm$ 0.277}\\
\bottomrule
\end{tabular}
\label{tab:transform}
\end{small}
\end{table}
To frame the LGCP as a Bayesian posterior sampling problem, we treat the Gaussian Process as the prior distribution and the Poisson process as the likelihood function:
\begin{equation}
\label{eq:LGCPBayes}
\pi(\mathbf{x}) = \mathcal{N}(\mathbf{x}; \mu, K), L(\mathbf{x}) =  \prod_{i}\exp(x_iy_i - \alpha e^{x_i}).
\end{equation}
\textbf{Logistic regression (type-2)}: Here we consider the Bayesian logistic regression with the prior distribution $\pi(\mathbf{x}) = \mathcal{N}(0,I)$, and the logistic regression model $ P(y_i) = \text{Bernoulli} (\text{sigmoid}(\mathbf{x}^T\cdot\mathbf{u}_i))$. The Bayesian inference of the logistic regression model is performed on the Ionosphere dataset with $D=35$ and the Sonar dataset with $D=61$.

\textbf{Latent space of Variational Autoencoder (type-2)}: In this experiment, we investigate sampling in the latent space of a pre-trained Variational Autoencoder (VAE) on the binary MNIST dataset. The posterior distribution in the latent space is denoted as the combination of a Gaussian prior of latent variable $\mathbf{z}$ and the decoder $p_\theta(\mathbf{x}\vert\mathbf{z})$.

\begin{table*}[!t]
\caption{$\log {\hat{\mathcal{Z}}}$ estimation of the Bayesian problems. As the ground-truth value is not known, SMC (with 1024 scales) results are considered as gold standard. Models that fall within the statistical margin of error relative to the gold-standard values are highlighted in bold font. }
\vskip 0.15in
\centering
\begin{small}
\begin{tabular}{c c c c c}
\toprule
Model  & LGCP ($D=1600$) & Ionosphere ($D=35$)& Sonar ($D=61$) & VAE ($D=30$)   \\
\midrule
SMC (1024) & 506.96 $\pm$ 0.24 &-111.61 $\pm$ 0.03 & -108.38 $\pm$ 0.02 & -110.16 $\pm$ 0.41 \\
\midrule
SMC (256) & \bf 506.77 $\pm$ 0.68 &\bf -111.62 $\pm$ 0.05 & \bf -108.39 $\pm$ 0.04 & \bf -110.20 $\pm$ 0.55\\
{LFIS}  & { 505.53 $\pm$ 0.95} & {\bf -111.60 $\pm$ 0.01} & {\bf -108.38 $\pm$ 0.01} & {\bf -109.99 $\pm$ 0.08} \\
DDS & 503.01 $\pm$ 0.77 & \bf-111.58 $\pm$ 0.12 & -108.92 $\pm$ 0.26 & \bf -110.02 $\pm$ 0.06 \\
PIS & 506.34 $\pm$ 0.63& -111.69 $\pm$ 0.16& -109.35 $\pm$ 0.74 &  -109.96 $\pm$ 0.09  \\
AFT &  505.96 $\pm$ 1.19 & -121.63 $\pm$ 16.37 & -104.79 $\pm$ 68.31& \bf -110.07 $\pm$ 0.36 \\
\bottomrule
\end{tabular}
\label{tab:bayes}
\end{small}
\end{table*}

\subsection{Results}

Tables \ref{tab:transform} and \ref{tab:bayes} show the performance of different methods on type-1 and type-2 problems respectively. We visualize the samples and weight distributions of type-1 problems in Fig.~\ref{fig:MG2D}. Comprehensive results of different experimental settings can be found in Appendix \ref{app:additionalDetails}. Our numerical results showed that LFIS is capable of generating competitive samples to existing methods. For two type-1 applications, we observed reasonable estimates of $\log\hat{\mathcal{Z}}$. Ground-truth (GT) samples can be generated for these two test problems, and the samples generated by LFIS have the lowest sample-to-GT-sample Wasserstein-2 distance. We remark that LFIS performs even better than the conventionally regarded gold-standard SMC with the same number of scales ($T=256$). {For Effective Sample Size (ESS) \citep{liuMonteCarloStrategies2008}, LFIS also shows competitive performance among all methods studied. Note that both SMC and AFT perform resampling when ESS is lower than a certain threshold (0.98 for SMC and 0.3 for AFT \citep{AFT}), while LFIS, DDS, PIS do not resample. Even without resampling, LFIS can still achieve high ESS comparable to SMC with a high resampling threshold.} The samples generated by LFIS also provide more uniform coverage over the different modes of the Gaussian mixture and the tails of funnel distribution compared to all other methods. In addition, the weight distributions of LFIS in these two problems are significantly narrower than other methods, explaining the higher ESS. For type-2 problems, LFIS is the only method delivering estimates of $\log \hat{\mathcal{Z}}$ that are consistent with the gold-standard SMC (with $T=1024$) on all problems.

\section{Discussion} \label{sec:discussion}

We first provide a discussion contrasting LFIS to existing methods and highlight the originality of LFIS. Among many existing methodologies for sampling unnormalized density functions, most closely related to LFIS are sampling by tempered transitions \citep{nealSamplingMultimodalDistributions1996}, AIS \citep{nealAnnealedImportanceSampling2001a}, and SMC \citep{SMC}), VI-NF \citep{VI-NF}, AFTMC \citep{AFT}, PIS \citep{zhang2021path}, CR-ACFMC \citep{matthewsContinualRepeatedAnnealed2022}, and DDS \citep{DDS}. Table \ref{tab:modelComplexity} provides a summary of the differences between LFIS and other existing methods. 

\begin{figure*}[!t] 
    \centering
    \includegraphics[width=0.99\textwidth]{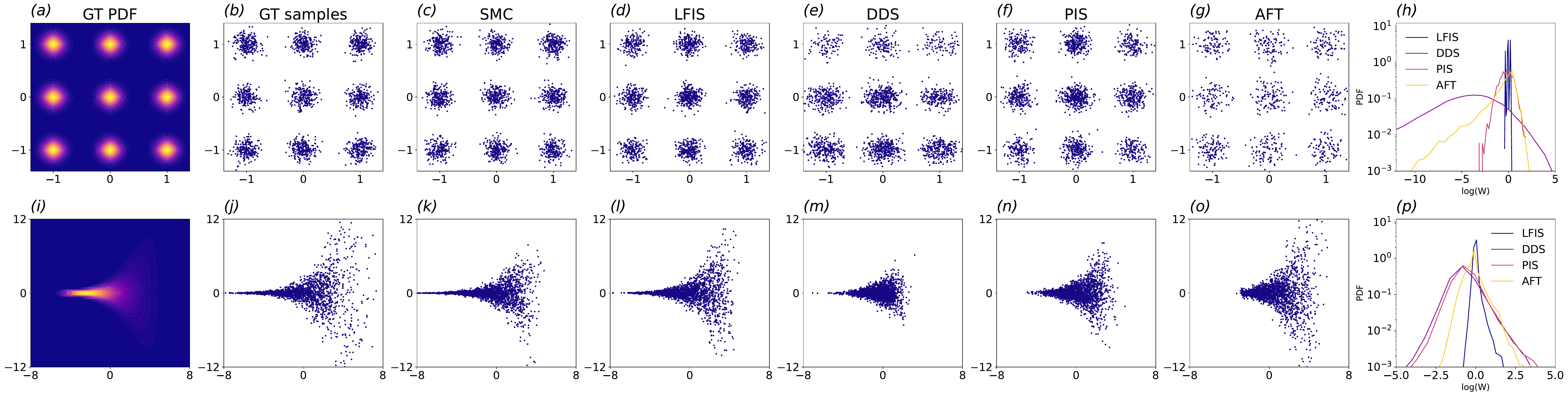}
    \caption{Sampling performance and sample weight distributions for the type-1 problems: (\textit{a-h}) 2-D Gaussian mixture and (\textit{i-p}) 10-D funnel distribution. Subfigures (\textit{a,i}) show the ground-truth PDF contours (marginalized 2-D contour for the funnel distribution). Subfigures (\textit{b-g} , \textit{j-o}) compare the generated samples from the ground-truth distribution and different sampling methods. For the funnel distribution, the samples are projected onto $(x_0,x_1)$ plane. Subfigures (\textit{h,p}) show the log-weight distributions for different sampling methods.} 
    \label{fig:MG2D}
\end{figure*}

LFIS shares the same spirit of a collection of MC methods \citep{nealSamplingMultimodalDistributions1996,nealAnnealedImportanceSampling2001a,SMC}, which all used a series of tempered densities for guiding the samplers converging to the target distribution. We adopted the specific form of the time-dependent distribution for type-1 applications from \citet{nealSamplingMultimodalDistributions1996} and \citet{nealAnnealedImportanceSampling2001a}, and for type-2 applications from \citet{SMC}. In addition, the weights derived by change-of-measure (see Appendix \ref{app:derivations}) align with the sequential importance sampling technique employed in these MC methods. The distinct difference between our proposed method to these MC-based methods is the transition dynamics between the prescribed series of densities: our proposition is a deterministic flow, in contrast to the stochastic Markov chains \citep{nealSamplingMultimodalDistributions1996,nealAnnealedImportanceSampling2001a} or particle filters \cite{SMC}. The performance of these MC-based methods critically depends on the choice of transition kernels, which are often an MCMC step, used for evolving samples between consecutive intermediate densities. This requires extensive engineering, i.e., tuning the meta parameters. In contrast, taking a flow-based model to evolve the samples allows us to derive the key equation \eqref{eq:consistent} and formulate an equation-based learning problem, which may streamline the modeling process.  

VI-NF \citep{VI-NF} uses a normalizing flow as the modeling distribution, which is parametrized by variational inference. LFIS shares many similar features of VI-NF when a Neural ODE \citep{neuralODE} is used as the flow. In this case, both our proposition and VI-NF draw samples from simple initial distributions ($\mu$ in our case; the base distribution in VI-NF) and use the NN-modeled flow to transport the samples to a future time. Both methods leverage the computable density function along the sample trajectory, a unique feature of flow-based models (see Eq.~\eqref{eq:logrhotheta-main}). The key difference between our proposition and VI-NF is the learning target. VI-NF is an end-to-end approach, that the loss function only depends on matching the final distribution at the end of the flow to the target distribution. In contrast, LFIS aims to capture $\tilde{\rho}_\ast(t)$, $t\in \left[0,1\right]$. 
We hypothesized that providing the whole evolution $\tilde{\rho}_\ast(t)$, instead of only the final $\tilde{\nu}=\tilde{\rho}_\ast(1)$, could improve the learning\footnote{As the same philosophy applies to AIS and SMC (analogous to LFIS) to simple MCMC-type inference (analogous to VI-NF).}. Moreover, VI-NF aims to directly optimize the reverse Kullback--Leibler (KL) divergence. In contrast, our approach first establishes the PDE \eqref{eq:consistent} for the \emph{optimal flow} which perfectly matches the evolution of the target distribution. Then, we parametrize the neural flow model by minimizing its discrepancy to the evolutionary equation, an approach similar to the Physics Informed Neural Networks \citep{RAISSI2019686,karniadakisPhysicsinformedMachineLearning2021}. Because VI-NF has been shown to perform worse than AFTMC \cite{AFT}, we did not include VI-NF in our quantitative comparison in Sec.~\ref{sec:numerical}.

The theoretical underpinning of LFIS, Eq.~\eqref{eq:consistent}, appeared in \citet{Jarzynsky08} and AFTMC \citep{AFT}. However, our proposition is significantly different from these studies. In the context of stochastic thermodynamics, \citet{Jarzynsky08} aimed to obtain the time-dependent density function for a prescribed time-dependent velocity field, contrasting our focus where we provide the unnormalized density function to solve for the velocity field. On the other hand, AFTMC sought to learn normalizing flows but entwined these flows and Monte Carlo kernels across the tempered scales. An ablation study in Appendix \ref{app:AFT-noMC} demonstrates that the bulk of learning in AFTMC was predominantly driven by the Monte Carlo kernels, aligning it more closely with SMC \cite{SMC} than a pure flow model like LFIS or VI-NF.  Without the normalizing flows, AFTMC is functionally identical to SMC. Given that the Monte Carlo kernels can be refined through optimization and that plain SMC competently handles the test problems (as observed in Appendix \ref{app:SMC}), the true advantage brought forth by the inclusion of normalizing flows in AFTMC remains nebulous. We remark that CR-AFTMC \citep{matthewsContinualRepeatedAnnealed2022}, similar to the construct of AFTMC but has some nuanced differences, also relies on the MC step. 
Gibbs Flow Approximation (GFA, \citet{hengGibbsFlowApproximate2021}) also proposed to leverage the Liouville equation for learning the velocity field, employing a Gibbs sampler instead of relying on Eq.~\eqref{eq:consistent} and its associated importance sampling. GFA also integrated Monte Carlo kernels in a manner akin to AFTMC. Nonetheless, it remains uncertain whether the dimension-by-dimension transportation framework of GFA is capable of addressing high-dimensional inference problems effectively.

LFIS is significantly different from other hybrid methods combining AIS and NF, for example,  \citet{NEURIPS2020_41d80bfc,NEURIPS2021_05f971b5,NEURIPS2021_a1a609f1,pmlr-v139-thin21a,NEURIPS2022_86b7128e,pmlr-v206-geffner23a}. This class of models also rely heavily on Monte Carlo sampling and thus requires special engineering i.e.~fine-tuning the Monte Carlo kernel on a case-by-case problem setting. 

PIS \citep{zhang2021path} and DDS \citep{DDS} are diffusion-based models that share many similar features. PIS learns the drift of an It\^o process and DDS learns the time-dependent score function. Both PIS and DDS parametrize the neural network in the path space, by minimizing the reverse KL divergence from a reference process\footnote{We adopted the nomenclature of DDS, which used ``reference process'' (PIS used the term ``prior uncontrolled process''). A subtle difference between PIS and DDS is the choice of this reference process: PIS uses the standard Wiener process, while DDS uses either over- or under-damped Ornstein--Uhlenbeck process.} to a modeling process, where the importance sampling is done through a change of measure via the Girsanov theorem \citep{protter2005stochastic}. PIS and DDS both are specific examples of a wider class of Schr\"odinger Bridge problem \citep{pavonStochasticControlNonequilibrium1989,daipraStochasticControlApproach1991,NEURIPS2021_940392f5}. Similar to VI-NF, PIS and DDS are both end-to-end and only the terminal distribution $\nu$ influences the learning. The performance of this class of models critically depends on the choice of the reference process and how it covers the target distribution. LFIS is closely related to the path-based philosophy of PIS and DDS, but it aims to learn the entire path $\rho_\ast(t)$, $t\in \left[0,1\right]$. PIS, DDS, and LFIS all estimate the weights of the samples, which can be understood as an integration of the deviation from some optimality along the paths.
As a deterministic flow, the path-measure of a particular trajectory with LFIS is always singular ($\delta$ distributions) and Girsanov transformation is not applicable. Consequently, the mathematical derivation of the importance sampler for LFIS is original and significantly different from the ones in PIS and DDS. 
We remark that methods using change of measure as a correction of imperfect computation exist, in addition to AFTMC, PIS, and DDS, see \cite{chorinImplicitSamplingParticle2009,morzfeldParameterEstimationImplicit2015a,goodmanSmallNoiseAnalysis2016,leachSymmetrizedImportanceSamplers2018}. However, it is beyond the authors' knowledge that the accumulated error of Eq.~\eqref{eq:consistent} along the trajectory plays a role as the sample weight for a deterministically transported dynamical system. 
Importance sampling by quantifying the accrued error along the trajectories significantly improved the accuracy of the model (see Appendix \ref{app:correction}). 
As such, we advocate the novelty of the theoretical construct of LFIS.
LFIS is more efficient in sampling than PIS without the need to generate random sample paths, but similar to DDS as it utilizes probability flow ODE \cite{songScoreBasedGenerativeModeling2021a}, which is a flow model and can be parametrized by our proposed method\footnote{The drift of the probability flow has a noise-induced term that depends on the score function already, so Eq.~\eqref{eq:consistent} for probability flow ODE will have an even higher-order derivative $\nabla_x^2 \log \rho_\ast(t)$.}.

LFIS is a sequential importance sampler and it accrues error as the samples evolve. This leads to an expansion in the spread of the weight distribution, consequently reducing the corresponding ESS of the sampler and diminishing its overall quality. This phenomenon is commonly observed in most sequential importance samplers, e.g., AIS, SMC, AFTMC, and CR-AFTMC. To mitigate this issue, one common approach is to perform resampling of the samples as in SMC, AFTMC and CR-AFTMC. In our experiments, we did not observe the weight distribution of LFIS deteriorating to the extent where resampling was deemed necessary. All the results in this manuscript are without resampling, although the incorporation of resampling into LFIS remains a viable option. Conducting LFIS without resampling can be viewed as a stringent test when compared to methods like SMC, AFTMC, or CR-AFTMC. Nonetheless, we still observed narrower weight distributions in LFIS, when compared to other methods, as illustrated in Figure \ref{fig:MG2D} (h) and (p) and also in type-2 problems (data not shown).

We conclude the manuscript by listing the potential limitations of LFIS. (1) LFIS cannot handle non-differentiable density function $\tilde{\nu}(x)$. For example, bounded uniform priors. Theoretically, these singularities can be handled by inserting sources at the discontinuities, a direction that merits future developments. A practical solution to address the challenge is to consider the non-differentiable density function as the limit of differentiable $\tilde{\rho}_\ast(x,t)$. For example, using sigmoid functions $\tilde{\nu}(x,t) = \sigma(tx/(1-t))$ which converge to step functions $\tilde{\nu}(x)={\Theta}(x)$. (2) When the flow satisfying Eq.~\eqref{eq:consistent} is too complex, LFIS' approach may require a more expressive NN than end-to-end DDS and PIS require. This can result in a more resource-intensive training process, but we speculate that it might offer a tradeoff in terms of improved accuracy. (3) LFIS is more memory-demanding as it requires $\nabla_x S_\ast$ in the equation, and cross terms like $\partial_\theta \partial_x v_\theta(x,t)$ in the optimization step. These terms are currently evaluated by memory-demanding \texttt{autograd}. (4) As an integrator and with the currently adopted explicit scheme, LFIS does not perform well when $T$ is small due to the error induced by a finite time-step (see Fig.~\ref{fig:steps}). This may result in a higher training cost. Optimizing LFIS using higher-order integration schemes and an interpolation of the neural flow along the time domain merits future research. 

\section*{Code availability}
The code for LFIS and the results of numerical experiments have been deposited at \url{https://github.com/lanl/LFIS}.

\section*{Acknowledgments}
The authors acknowledge continual support from Laboratory Directed Research and Development (LDRD). YT was supported by LDRD project ``Accelerated Dynamics Across Computational and Physical Scales'' (220063DR), NP was supported by LDRD project ``Learning Uncertainties In Coupled-Physics Models via Operator Theory'' (20230254ER), and YTL was supported by LDRD project ``Diffusion Modeling with Physical Constraints for Scientific Data'' (20240074ER). YTL sincerely thanks Prof.~A.~Doucet for several insightful email exchanges that partially inspired this work.

\section*{Impact Statement} This manuscript presents work whose goal is to advance the field of statistical sampling and machine learning. There are many potential societal consequences of our work, none of which we feel must be specifically highlighted here.

\newpage
\bibliography{ref}

\begin{thebibliography}{64}
\providecommand{\natexlab}[1]{#1}
\providecommand{\url}[1]{\texttt{#1}}
\expandafter\ifx\csname urlstyle\endcsname\relax
  \providecommand{\doi}[1]{doi: #1}\else
  \providecommand{\doi}{doi: \begingroup \urlstyle{rm}\Url}\fi

\bibitem[Arbel et~al.(2021{\natexlab{a}})Arbel, Matthews, and Doucet]{AFT}
Arbel, M., Matthews, A., and Doucet, A.
\newblock Annealed flow transport monte carlo.
\newblock In \emph{Proceedings of the 38th International Conference on Machine Learning}, volume 139 of \emph{Proceedings of Machine Learning Research}, pp.\  318--330. {PMLR}, July 2021{\natexlab{a}}.

\bibitem[Arbel et~al.(2021{\natexlab{b}})Arbel, Matthews, and Doucet]{AFTrepo}
Arbel, M., Matthews, A., and Doucet, A.
\newblock {Github repository of Annealed Flow Transport Monte Carlo Sampler}.
\newblock \url{https://github.com/google-deepmind/annealed_flow_transport}, 2021{\natexlab{b}}.
\newblock [Online; accessed 4-Jan-2024].

\bibitem[Bhatia et~al.(2012)Bhatia, Norgard, Pascucci, and Bremer]{bhatia2012helmholtz}
Bhatia, H., Norgard, G., Pascucci, V., and Bremer, P.-T.
\newblock The {Helmholtz-Hodge} decomposition—a survey.
\newblock \emph{IEEE Transactions on visualization and computer graphics}, 19\penalty0 (8):\penalty0 1386--1404, 2012.

\bibitem[Bierkens et~al.(2019)Bierkens, Fearnhead, and Roberts]{bierkensZigZagProcessSuperefficient2019}
Bierkens, J., Fearnhead, P., and Roberts, G.
\newblock The {{Zig-Zag}} process and super-efficient sampling for {{Bayesian}} analysis of big data.
\newblock \emph{The Annals of Statistics}, 47\penalty0 (3), June 2019.
\newblock ISSN 0090-5364.
\newblock \doi{10.1214/18-AOS1715}.

\bibitem[Bonneel et~al.()Bonneel, Rabin, Peyré, and Pfister]{bonneelSlicedRadonWasserstein2015}
Bonneel, N., Rabin, J., Peyré, G., and Pfister, H.
\newblock Sliced and {{Radon Wasserstein Barycenters}} of {{measures}}.
\newblock 51\penalty0 (1):\penalty0 22--45.
\newblock ISSN 1573-7683.
\newblock \doi{10.1007/s10851-014-0506-3}.

\bibitem[Chen et~al.(2018)Chen, Rubanova, Bettencourt, and Duvenaud]{neuralODE}
Chen, R. T.~Q., Rubanova, Y., Bettencourt, J., and Duvenaud, D.~K.
\newblock Neural ordinary differential equations.
\newblock In \emph{Advances in Neural Information Processing Systems}, volume~31, 2018.

\bibitem[Chorin \& Tu(2009)Chorin and Tu]{chorinImplicitSamplingParticle2009}
Chorin, A.~J. and Tu, X.
\newblock Implicit sampling for particle filters.
\newblock \emph{Proceedings of the National Academy of Sciences}, 106\penalty0 (41):\penalty0 17249--17254, October 2009.
\newblock ISSN 0027-8424, 1091-6490.
\newblock \doi{10.1073/pnas.0909196106}.

\bibitem[Dai~Pra(1991)]{daipraStochasticControlApproach1991}
Dai~Pra, P.
\newblock A stochastic control approach to reciprocal diffusion processes.
\newblock \emph{Applied Mathematics and Optimization}, 23\penalty0 (1):\penalty0 313--329, January 1991.
\newblock ISSN 1432-0606.
\newblock \doi{10.1007/BF01442404}.

\bibitem[De~Bortoli et~al.(2021)De~Bortoli, Thornton, Heng, and Doucet]{NEURIPS2021_940392f5}
De~Bortoli, V., Thornton, J., Heng, J., and Doucet, A.
\newblock Diffusion {S}chr\"odinger {B}ridge with applications to score-based generative modeling.
\newblock In \emph{Advances in Neural Information Processing Systems}, volume~34, pp.\  17695--17709, 2021.

\bibitem[Del~Moral et~al.(2006)Del~Moral, Doucet, and Jasra]{SMC}
Del~Moral, P., Doucet, A., and Jasra, A.
\newblock Sequential {{Monte Carlo Samplers}}.
\newblock \emph{Journal of the Royal Statistical Society. Series B (Statistical Methodology)}, 68\penalty0 (3):\penalty0 411--436, 2006.
\newblock ISSN 13697412, 14679868.

\bibitem[Doucet et~al.(2022)Doucet, Grathwohl, Matthews, and Strathmann]{NEURIPS2022_86b7128e}
Doucet, A., Grathwohl, W., Matthews, A.~G., and Strathmann, H.
\newblock Score-based diffusion meets annealed importance sampling.
\newblock In \emph{Advances in Neural Information Processing Systems}, volume~35, pp.\  21482--21494, 2022.

\bibitem[Duane et~al.(1987)Duane, Kennedy, Pendleton, and Roweth]{duaneHybridMonteCarlo1987}
Duane, S., Kennedy, A., Pendleton, B.~J., and Roweth, D.
\newblock Hybrid {{Monte Carlo}}.
\newblock \emph{Physics Letters B}, 195\penalty0 (2):\penalty0 216--222, September 1987.
\newblock ISSN 0370-2693.
\newblock \doi{10.1016/0370-2693(87)91197-X}.

\bibitem[Evans(2022)]{evans2022partial}
Evans, L.~C.
\newblock \emph{Partial differential equations}, volume~19.
\newblock American Mathematical Society, 2022.

\bibitem[Faulkner \& Livingstone(2023)Faulkner and Livingstone]{faulknerSamplingAlgorithmsStatistical2023}
Faulkner, M.~F. and Livingstone, S.
\newblock Sampling algorithms in statistical physics: A guide for statistics and machine learning, June 2023.

\bibitem[Gabri{\'e} et~al.(2022)Gabri{\'e}, Rotskoff, and {Vanden-Eijnden}]{gabrieAdaptiveMonteCarlo2022}
Gabri{\'e}, M., Rotskoff, G.~M., and {Vanden-Eijnden}, E.
\newblock Adaptive {{Monte Carlo}} augmented with {Normalizing Flows}.
\newblock \emph{Proceedings of the National Academy of Sciences}, 119\penalty0 (10):\penalty0 e2109420119, March 2022.
\newblock ISSN 0027-8424, 1091-6490.
\newblock \doi{10.1073/pnas.2109420119}.

\bibitem[Geffner \& Domke(2021)Geffner and Domke]{NEURIPS2021_05f971b5}
Geffner, T. and Domke, J.
\newblock {{MCMC}} variational inference via uncorrected {H}amiltonian annealing.
\newblock In \emph{Advances in Neural Information Processing Systems}, volume~34, pp.\  639--651, 2021.

\bibitem[Geffner \& Domke(2023)Geffner and Domke]{pmlr-v206-geffner23a}
Geffner, T. and Domke, J.
\newblock Langevin {D}iffusion {V}ariational {I}nference.
\newblock In \emph{Proceedings of the 26th International Conference on Artificial Intelligence and Statistics}, volume 206 of \emph{Proceedings of Machine Learning Research}, pp.\  576--593. {PMLR}, April 2023.

\bibitem[Gelman \& Meng(1998)Gelman and Meng]{gelmanSimulatingNormalizingConstants1998}
Gelman, A. and Meng, X.-L.
\newblock Simulating normalizing constants: From importance sampling to {Bridge Sampling} to {Path Sampling}.
\newblock \emph{Statistical Science}, 13\penalty0 (2), May 1998.
\newblock ISSN 0883-4237.
\newblock \doi{10.1214/ss/1028905934}.

\bibitem[Gerlich(1973)]{gerlich1973VerallgemeinerteLiouvilleGleichung}
Gerlich, G.
\newblock Die verallgemeinerte {{Liouville}}-{{Gleichung}}.
\newblock \emph{Physica}, 69\penalty0 (2):\penalty0 458--466, November 1973.
\newblock ISSN 0031-8914.
\newblock \doi{10.1016/0031-8914(73)90083-9}.

\bibitem[Geweke(1989)]{geweke1989bayesian}
Geweke, J.
\newblock Bayesian inference in econometric models using monte carlo integration.
\newblock \emph{Econometrica: Journal of the Econometric Society}, pp.\  1317--1339, 1989.

\bibitem[Goodman et~al.(2016)Goodman, Lin, and Morzfeld]{goodmanSmallNoiseAnalysis2016}
Goodman, J., Lin, K.~K., and Morzfeld, M.
\newblock Small-{{Noise Analysis}} and {{Symmetrization}} of {{Implicit Monte Carlo Samplers}}.
\newblock \emph{Communications on Pure and Applied Mathematics}, 69\penalty0 (10):\penalty0 1924--1951, October 2016.
\newblock ISSN 0010-3640, 1097-0312.
\newblock \doi{10.1002/cpa.21592}.

\bibitem[Green(1995)]{RJMCMC}
Green, P.~J.
\newblock {Reversible jump {M}arkov chain {M}onte {C}arlo computation and Bayesian model determination}.
\newblock \emph{Biometrika}, 82\penalty0 (4):\penalty0 711--732, 12 1995.
\newblock ISSN 0006-3444.
\newblock \doi{10.1093/biomet/82.4.711}.

\bibitem[Heng et~al.(2021)Heng, Doucet, and Pokern]{hengGibbsFlowApproximate2021}
Heng, J., Doucet, A., and Pokern, Y.
\newblock Gibbs flow for approximate transport with applications to bayesian computation.
\newblock 83\penalty0 (1):\penalty0 156--187, 2021.
\newblock ISSN 1369-7412.
\newblock \doi{10.1111/rssb.12404}.

\bibitem[H{\'e}nin et~al.(2022)H{\'e}nin, Leli{\`e}vre, Shirts, Valsson, and Delemotte]{heninEnhancedSamplingMethods2022}
H{\'e}nin, J., Leli{\`e}vre, T., Shirts, M.~R., Valsson, O., and Delemotte, L.
\newblock Enhanced sampling methods for molecular dynamics simulations [{{Article}} v1.0].
\newblock \emph{Living Journal of Computational Molecular Science}, 4\penalty0 (1):\penalty0 1583, December 2022.
\newblock \doi{10.33011/livecoms.4.1.1583}.

\bibitem[Hines(2015)]{HINES20152103}
Hines, K.~E.
\newblock A primer on {B}ayesian inference for biophysical systems.
\newblock \emph{Biophysical Journal}, 108\penalty0 (9):\penalty0 2103--2113, 2015.
\newblock ISSN 0006-3495.
\newblock \doi{10.1016/j.bpj.2015.03.042}.

\bibitem[Jackman(2000)]{jackmanEstimationInferenceBayesian2000}
Jackman, S.
\newblock Estimation and {{Inference}} via {{Bayesian Simulation}}: {{An Introduction}} to {{Markov Chain Monte Carlo}}.
\newblock \emph{American Journal of Political Science}, 44\penalty0 (2):\penalty0 375, April 2000.
\newblock ISSN 00925853.
\newblock \doi{10.2307/2669318}.

\bibitem[Karniadakis et~al.(2021)Karniadakis, Kevrekidis, Lu, Perdikaris, Wang, and Yang]{karniadakisPhysicsinformedMachineLearning2021}
Karniadakis, G.~E., Kevrekidis, I.~G., Lu, L., Perdikaris, P., Wang, S., and Yang, L.
\newblock Physics-informed machine learning.
\newblock \emph{Nature Reviews Physics}, 3\penalty0 (6):\penalty0 422--440, June 2021.
\newblock ISSN 2522-5820.
\newblock \doi{10.1038/s42254-021-00314-5}.

\bibitem[Kass \& Raftery(1995)Kass and Raftery]{BayesFactors2}
Kass, R.~E. and Raftery, A.~E.
\newblock {B}ayes factors.
\newblock \emph{Journal of the American Statistical Association}, 90\penalty0 (430):\penalty0 773--795, 1995.
\newblock \doi{10.1080/01621459.1995.10476572}.

\bibitem[Larsson \& Thom{\'e}e(2003)Larsson and Thom{\'e}e]{larsson2003partial}
Larsson, S. and Thom{\'e}e, V.
\newblock \emph{Partial differential equations with numerical methods}, volume~45.
\newblock Springer, 2003.

\bibitem[Leach et~al.(2018)Leach, Lin, and Morzfeld]{leachSymmetrizedImportanceSamplers2018}
Leach, A., Lin, K.~K., and Morzfeld, M.
\newblock Symmetrized importance samplers for stochastic differential equations.
\newblock \emph{Communications in Applied Mathematics and Computational Science}, 13\penalty0 (2):\penalty0 215--241, June 2018.
\newblock ISSN 2157-5452, 1559-3940.
\newblock \doi{10.2140/camcos.2018.13.215}.

\bibitem[Li et~al.(2019)Li, Holbrook, Shahbaba, and Baldi]{liNeuralNetworkGradient2019}
Li, L., Holbrook, A., Shahbaba, B., and Baldi, P.
\newblock Neural network gradient {{Hamiltonian Monte Carlo}}.
\newblock \emph{Computational Statistics}, 34\penalty0 (1):\penalty0 281--299, March 2019.
\newblock ISSN 0943-4062, 1613-9658.
\newblock \doi{10.1007/s00180-018-00861-z}.

\bibitem[Lipman et~al.(2023)Lipman, Chen, Ben-Hamu, Nickel, and Le]{lipman2023flow}
Lipman, Y., Chen, R. T.~Q., Ben-Hamu, H., Nickel, M., and Le, M.
\newblock Flow matching for generative modeling.
\newblock In \emph{The Eleventh International Conference on Learning Representations}, 2023.
\newblock URL \url{https://openreview.net/forum?id=PqvMRDCJT9t}.

\bibitem[Liu(2008)]{liuMonteCarloStrategies2008}
Liu, J.~S.
\newblock \emph{Monte {{Carlo}} Strategies in Scientific Computing}.
\newblock Springer Series in Statistics. {Springer}, {New York, NY}, 2. ed edition, 2008.
\newblock ISBN 978-0-387-76369-9.

\bibitem[Llorente et~al.(2023)Llorente, Martino, Delgado, and {L{\'o}pez-Santiago}]{llorenteMarginalLikelihoodComputation2023}
Llorente, F., Martino, L., Delgado, D., and {L{\'o}pez-Santiago}, J.
\newblock Marginal likelihood computation for model selection and hypothesis testing: {{An}} extensive review.
\newblock \emph{SIAM Review}, 65\penalty0 (1):\penalty0 3--58, 2023.
\newblock \doi{10.1137/20M1310849}.

\bibitem[Matthews et~al.(2022)Matthews, Arbel, Rezende, and Doucet]{matthewsContinualRepeatedAnnealed2022}
Matthews, A., Arbel, M., Rezende, D.~J., and Doucet, A.
\newblock Continual repeated annealed flow transport {{Monte Carlo}}.
\newblock In \emph{Proceedings of the 39th International Conference on Machine Learning}, volume 162 of \emph{Proceedings of Machine Learning Research}, pp.\  15196--15219. {PMLR}, July 2022.

\bibitem[McCartan \& Imai(2023)McCartan and Imai]{mccartanSequentialMonteCarlo2023}
McCartan, C. and Imai, K.
\newblock Sequential {{Monte Carlo}} for sampling balanced and compact redistricting plans.
\newblock \emph{The Annals of Applied Statistics}, 17\penalty0 (4), December 2023.
\newblock ISSN 1932-6157.
\newblock \doi{10.1214/23-AOAS1763}.

\bibitem[Metropolis et~al.(1953)Metropolis, Rosenbluth, Rosenbluth, Teller, and Teller]{metropolisEquationStateCalculations1953}
Metropolis, N., Rosenbluth, A.~W., Rosenbluth, M.~N., Teller, A.~H., and Teller, E.
\newblock Equation of state calculations by fast computing machines.
\newblock \emph{The Journal of Chemical Physics}, 21\penalty0 (6):\penalty0 1087--1092, 1953.
\newblock ISSN 0021-9606.
\newblock \doi{10.1063/1.1699114}.

\bibitem[M{\o}ller et~al.(1998)M{\o}ller, Syversveen, and Waagepetersen]{LGCP}
M{\o}ller, J., Syversveen, A.~R., and Waagepetersen, R.~P.
\newblock Log {G}aussian {C}ox {P}rocesses.
\newblock \emph{Scandinavian Journal of Statistics}, 25\penalty0 (3):\penalty0 451--482, 1998.

\bibitem[Morzfeld et~al.(2015)Morzfeld, Tu, Wilkening, and Chorin]{morzfeldParameterEstimationImplicit2015a}
Morzfeld, M., Tu, X., Wilkening, J., and Chorin, A.
\newblock Parameter estimation by implicit sampling.
\newblock \emph{Communications in Applied Mathematics and Computational Science}, 10\penalty0 (2):\penalty0 205--225, September 2015.
\newblock ISSN 2157-5452, 1559-3940.
\newblock \doi{10.2140/camcos.2015.10.205}.

\bibitem[Neal(1996)]{nealSamplingMultimodalDistributions1996}
Neal, R.~M.
\newblock Sampling from multimodal distributions using tempered transitions.
\newblock \emph{Statistics and Computing}, 6\penalty0 (4):\penalty0 353--366, December 1996.
\newblock ISSN 1573-1375.
\newblock \doi{10.1007/BF00143556}.

\bibitem[Neal(2001)]{nealAnnealedImportanceSampling2001a}
Neal, R.~M.
\newblock Annealed importance sampling.
\newblock \emph{Statistics and Computing}, 11\penalty0 (2):\penalty0 125--139, April 2001.
\newblock ISSN 1573-1375.
\newblock \doi{10.1023/A:1008923215028}.

\bibitem[Neal(2003)]{nealSliceSampling2003}
Neal, R.~M.
\newblock Slice sampling.
\newblock \emph{The Annals of Statistics}, 31\penalty0 (3):\penalty0 705--767, June 2003.
\newblock ISSN 0090-5364, 2168-8966.
\newblock \doi{10.1214/aos/1056562461}.

\bibitem[Neal(2012)]{nealMCMCUsingHamiltonian2012}
Neal, R.~M.
\newblock {{MCMC}} using {{Hamiltonian}} dynamics.
\newblock \emph{arXiv:1206.1901 [physics, stat]}, June 2012.
\newblock \doi{10.1201/b10905}.

\bibitem[Pajor(2017)]{pajorEstimatingMarginalLikelihood2017}
Pajor, A.
\newblock Estimating the {{Marginal Likelihood Using}} the {{Arithmetic Mean Identity}}.
\newblock \emph{Bayesian Analysis}, 12\penalty0 (1):\penalty0 261--287, 2017.
\newblock ISSN 1936-0975.
\newblock \doi{10.1214/16-BA1001}.

\bibitem[Pavon(1989)]{pavonStochasticControlNonequilibrium1989}
Pavon, M.
\newblock Stochastic control and nonequilibrium thermodynamical systems.
\newblock \emph{Applied Mathematics and Optimization}, 19\penalty0 (1):\penalty0 187--202, January 1989.
\newblock ISSN 1432-0606.
\newblock \doi{10.1007/BF01448198}.

\bibitem[Protter \& Protter(2005)Protter and Protter]{protter2005stochastic}
Protter, P.~E. and Protter, P.~E.
\newblock \emph{Stochastic differential equations}.
\newblock Springer, 2005.

\bibitem[Raissi et~al.(2019)Raissi, Perdikaris, and Karniadakis]{RAISSI2019686}
Raissi, M., Perdikaris, P., and Karniadakis, G.
\newblock Physics-informed neural networks: {{A}} deep learning framework for solving forward and inverse problems involving nonlinear partial differential equations.
\newblock \emph{Journal of Computational Physics}, 378:\penalty0 686--707, 2019.
\newblock ISSN 0021-9991.
\newblock \doi{10.1016/j.jcp.2018.10.045}.

\bibitem[Resnick(2019)]{resnick2019probability}
Resnick, S.
\newblock \emph{A probability path}.
\newblock Springer, 2019.

\bibitem[Rezende \& Mohamed(2015)Rezende and Mohamed]{VI-NF}
Rezende, D. and Mohamed, S.
\newblock Variational inference with {N}ormalizing {F}lows.
\newblock In \emph{Proceedings of the 32nd International Conference on Machine Learning}, volume~37 of \emph{Proceedings of Machine Learning Research}, pp.\  1530--1538, Lille, France, 07--09 Jul 2015. PMLR.

\bibitem[Robert \& Wraith(2009)Robert and Wraith]{robertComputationalMethodsBayesian2009}
Robert, C.~P. and Wraith, D.
\newblock Computational methods for {{Bayesian}} model choice.
\newblock \emph{AIP Conference Proceedings}, 1193\penalty0 (1):\penalty0 251, 2009.
\newblock \doi{10.1063/1.3275622}.

\bibitem[Sharma(2017)]{sharmaMarkovChainMonte2017}
Sharma, S.
\newblock Markov chain {M}onte {C}arlo methods for {B}ayesian data analysis in astronomy.
\newblock \emph{Annual Review of Astronomy and Astrophysics}, 55\penalty0 (1):\penalty0 213--259, 2017.
\newblock \doi{10.1146/annurev-astro-082214-122339}.

\bibitem[Song et~al.()Song, Sohl-Dickstein, Kingma, Kumar, Ermon, and Poole]{songScoreBasedGenerativeModeling2021a}
Song, Y., Sohl-Dickstein, J., Kingma, D.~P., Kumar, A., Ermon, S., and Poole, B.
\newblock Score-{{Based Generative Modeling}} through {{Stochastic Differential Equations}}.
\newblock Comment: ICLR 2021 (Oral).

\bibitem[Thin et~al.(2021)Thin, Kotelevskii, Doucet, Durmus, Moulines, and Panov]{pmlr-v139-thin21a}
Thin, A., Kotelevskii, N., Doucet, A., Durmus, A., Moulines, E., and Panov, M.
\newblock Monte {C}arlo variational auto-encoders.
\newblock In \emph{Proceedings of the 38th International Conference on Machine Learning}, volume 139 of \emph{Proceedings of Machine Learning Research}, pp.\  10247--10257. {PMLR}, July 2021.

\bibitem[Tokdar \& Kass(2010)Tokdar and Kass]{tokdar2010importance}
Tokdar, S.~T. and Kass, R.~E.
\newblock Importance sampling: a review.
\newblock \emph{Wiley Interdisciplinary Reviews: Computational Statistics}, 2\penalty0 (1):\penalty0 54--60, 2010.

\bibitem[Tong et~al.(2024)Tong, FATRAS, Malkin, Huguet, Zhang, Rector-Brooks, Wolf, and Bengio]{tong2024improving}
Tong, A., FATRAS, K., Malkin, N., Huguet, G., Zhang, Y., Rector-Brooks, J., Wolf, G., and Bengio, Y.
\newblock Improving and generalizing flow-based generative models with minibatch optimal transport.
\newblock \emph{Transactions on Machine Learning Research}, 2024.
\newblock ISSN 2835-8856.
\newblock URL \url{https://openreview.net/forum?id=CD9Snc73AW}.
\newblock Expert Certification.

\bibitem[Vaikuntanathan \& Jarzynski(2008)Vaikuntanathan and Jarzynski]{Jarzynsky08}
Vaikuntanathan, S. and Jarzynski, C.
\newblock Escorted free energy simulations: Improving convergence by reducing dissipation.
\newblock \emph{Phys. Rev. Lett.}, 100:\penalty0 190601, May 2008.
\newblock \doi{10.1103/PhysRevLett.100.190601}.

\bibitem[Vargas et~al.(2023{\natexlab{a}})Vargas, Grathwohl, and Doucet]{DDS}
Vargas, F., Grathwohl, W.~S., and Doucet, A.
\newblock Denoising {D}iffusion {S}amplers.
\newblock In \emph{The Eleventh International Conference on Learning Representations, {ICLR} 2023, Kigali, Rwanda, May 1-5, 2023}. OpenReview.net, 2023{\natexlab{a}}.

\bibitem[Vargas et~al.(2023{\natexlab{b}})Vargas, Grathwohl, and Doucet]{DDSrepo}
Vargas, F., Grathwohl, W.~S., and Doucet, A.
\newblock {Github repository of Denoising Diffusion Samplers}.
\newblock \url{https://github.com/franciscovargas/denoising_diffusion_samplers}, 2023{\natexlab{b}}.
\newblock [Online; accessed 4-Jan-2024].

\bibitem[Wainwright \& Jordan(2008)Wainwright and Jordan]{wainwrightGraphicalModelsExponential2008}
Wainwright, M.~J. and Jordan, M.~I.
\newblock Graphical models, exponential families, and variational inference.
\newblock \emph{Foundations and Trends{\textregistered} in Machine Learning}, 1\penalty0 (1{\textendash}2):\penalty0 1--305, 2008.
\newblock ISSN 1935-8237.
\newblock \doi{10.1561/2200000001}.

\bibitem[Wilkinson(2007)]{wilkinsonBayesianMethodsBioinformatics2007}
Wilkinson, D.~J.
\newblock Bayesian methods in bioinformatics and computational systems biology.
\newblock \emph{Briefings in Bioinformatics}, 8\penalty0 (2):\penalty0 109--116, April 2007.
\newblock ISSN 1467-5463.
\newblock \doi{10.1093/bib/bbm007}.

\bibitem[Wu et~al.(2020)Wu, K{\"o}hler, and Noe]{NEURIPS2020_41d80bfc}
Wu, H., K{\"o}hler, J., and Noe, F.
\newblock Stochastic normalizing flows.
\newblock In \emph{Advances in Neural Information Processing Systems}, volume~33, pp.\  5933--5944, 2020.

\bibitem[Zellner(1988)]{zellnerOptimalInformationProcessing1988}
Zellner, A.
\newblock Optimal information processing and {B}ayes's theorem.
\newblock \emph{The American Statistician}, 42\penalty0 (4):\penalty0 278--280, 1988.
\newblock \doi{10.1080/00031305.1988.10475585}.

\bibitem[Zhang et~al.(2021)Zhang, Hsu, Li, Finn, and Grosse]{NEURIPS2021_a1a609f1}
Zhang, G., Hsu, K., Li, J., Finn, C., and Grosse, R.~B.
\newblock Differentiable annealed importance sampling and the perils of gradient noise.
\newblock In \emph{Advances in Neural Information Processing Systems}, volume~34, pp.\  19398--19410, 2021.

\bibitem[Zhang \& Chen(2022)Zhang and Chen]{zhang2021path}
Zhang, Q. and Chen, Y.
\newblock Path integral sampler: A stochastic control approach for sampling.
\newblock In \emph{International Conference on Learning Representations}, 2022.

\end{thebibliography}

\bibliographystyle{icml2024}

\newpage
\onecolumn
\appendix
\section{Existence of Velocity Field} \label{app:existenceProof}

We prove theorem~\ref{thm:existence_vel_field}. We want to state that uniqueness is not guaranteed, however the existence of the velocity field in Eq.~\eqref{eq:consistent} \emph{only depends} on the condition that $\nabla_x\cdot S_{\ast}\left(x,t\right)$ is bounded for all $t$ \emph{almost everywhere}. Our proof shows that many solutions can exist, i.e., non-uniqueness. Our proposed Liouville Flow Importance Sampler models the velocity field using a neural network that satisfies Eq.~\eqref{eq:consistent} in a least squares sense, nonetheless the existence of the velocity field rests on a concrete mathematical foundation based on the Helmholtz--Hodge decomposition as shown below.

\begin{proof}{Theorem ~\ref{thm:existence_vel_field}.}
We first show existence.
\begin{subequations}
Let $t\in [0,1]$. By Helmholtz--Hodge decomposition (see the survey in~\cite{bhatia2012helmholtz}), we can decompose the velocity field ${v}(x,t)$ as
\begin{equation}
    v(x,t) = \nabla_x\psi(x;t) + u(x;t),
\end{equation}
where $u(x;t)$ is divergence free, that is $\nabla_x\cdot u = 0$.
Plugging this decomposition into Eq.~\eqref{eq:consistent}, we obtained,
\begin{equation}\label{eq:flow_all}
  \nabla_x^2 \psi(x;t) + S_{\ast}(x,t)\cdot \nabla_x \psi(x;t) + S_{\ast}(x,t)\cdot u(x;t) =  -f(x,t).
\end{equation} 
Next, by adding and subtracting $c\psi(x;t)$ to the equation above,
\begin{equation}\label{eq:flow_all_2}
  \nabla_x^2 \psi(x;t) + S_{\ast}(x,t)\cdot \nabla_x \psi(x;t) + c\psi(x;t) + S_{\ast}(x,t)\cdot u(x;t) - c\psi(x;t) =  -f(x,t).
\end{equation} 
By demanding that $\psi$ and $u$ satisfy the following equations respectively,
\begin{equation}\label{eq:Flow_elliptic}
\nabla_x^2 \psi(x;t) + S_{\ast}(x,t)\cdot\nabla_x \psi(x;t) + c\psi(x;t) = -f(x,t); \quad \psi(x,t) = 0 \quad \text{on boundary},
\end{equation}
and
\begin{equation}
\label{eq:Flow_curl}
    {S}_{\ast}(x,t)\cdot u(x;t) = c\psi(x;t),
\end{equation}
we see that $v(x,t) = \nabla_x\psi(x;t) + u(x;t)$ satisfies Eq.~\eqref{eq:consistent}. Observe that the equation in~\eqref{eq:Flow_elliptic} is an \emph{elliptic partial differential equation} of the form
\begin{equation}\label{eq:elliptic_pde}
   -\nabla_x\cdot\left(a\nabla_x\psi\right) + b\cdot\nabla\psi + d\psi = f, 
\end{equation}
where $a = 1$, $b = -S_{\ast}$ and $d = -c$.
By the Lax--Milgram theorem (see chp. $6$ in~\cite{evans2022partial} for a general description and sec 3.5 in~\cite{larsson2003partial} for the particular case described by Eq.~\eqref{eq:elliptic_pde}), \emph{a unique} $\psi(x,t)$ \emph{exists} provided $d - \frac{1}{2}\nabla_x\cdot b \geq 0$ for all $x,t$. In other words, $2d \geq -\nabla_x\cdot S_{\ast}(x;t)$, that is $2c \leq \nabla_{x}\cdot S_{\ast}$. By our assumption, this is indeed true.

Observe that if $v_1(x,t)$ is a solution to Eq.~\eqref{eq:consistent} and let $\tilde{v}(x,t)$ be a vector field such that $\left[\nabla + S_{\ast}\right]\cdot\tilde{v} = 0$ then $v_2(x,t) = v_1(x,t) + \tilde{v}(x,t)$ is also a solution to Eq.~\eqref{eq:consistent}. As $\left[\nabla + S_{\ast}\right]\cdot\tilde{v} = 0$ only provides one constraint on the $D$-dimensional field $\tilde{v}$, for $D\ge 2$, there is no unique solution to Eq.~\eqref{eq:consistent}.

\end{subequations}    
\end{proof}

\section{Importance Sampling for Imperfect Neural Networks} \label{app:derivations}

Almost surely, a finite network cannot accurately model the velocity field that satisfies Eq.~\eqref{eq:consistent}. Let us denote the neural velocity field by $v_{\theta} (x,t)$, where $\theta$ stands for the weights of the neural network. In this scenario, we denote the induced density function by $\rho_\theta(x,t)$, which in general is not the target density function $\rho_\ast(x,t)$. Note that GLE \eqref{eq:Liouville} ensures that the trajectories of $N$ samples following the deterministic evolution \eqref{eq:ODE} as given by Eq.~\eqref{eq:sample_traj_nn} share the same statistics of $\rho_\theta(\cdot, t)$. That is, $x^{(i)}(t)\sim \rho_\theta(\cdot, t)$ whenever $x_0^{(i)}\sim \mu$. Our goal is to derive an equation that is analogous to Eq.~\eqref{eq:consistent} using variational inference (VI) for this imperfect ($\rho_\theta \ne \rho_\ast$) scenario. 

To describe an instantaneous equation for the velocity field, it suffices to consider performing variational inference at time $t+\dd t$, where the infinitesimal advanced time $\dd t\ll 1$. Using VI, we aim to minimize the reverse Kullback--Leibler divergence $\text{KL}\left(\rho_\theta\left(\cdot, t+\dd t\right) \Vert \rho_\ast \left(\cdot, t+\dd t\right) \right)$. As we will be using the sample trajectories, it is convenient to first quantify both $\log\rho_\theta(x(t), t)$ and $\log \rho_\ast(x(t),t)$ along the trajectory. Note that we are adopting the Lagrangian specification of the flow because we are computing the quantity of interests along the trajectory. Below, for brevity, we denote a sample trajectory $x^{(i)}(t)$ by the abbreviated $x(t)$. We first compute the instantaneous rate of log-densities along the trajectories: 
\begin{align}
    \frac{\dd}{\dd t} \log \rho_\theta(x(t), t) ={}& \partial_x \left[ \log\rho_\theta(x(t), t) \right] \cdot \frac{\dd x(t)}{\dd t} +  \partial_t \log\rho_\theta(x(t), t) \nonumber \\
    ={}& S_\theta(x(t), t) \cdot v_\theta(x(t),t) -\frac{1}{\rho(x(t),t)}\nabla_x \cdot \left(v_\theta(x(t), t) \rho_\theta (x(t), t)\right)  \quad \quad  \text{(By Eq.~\eqref{eq:Liouville})} \nonumber \\ 
    ={}& -\nabla_x \cdot v_\theta\left(x(t),t\right), \\
    \frac{\dd}{\dd t} \log \rho_\ast(x(t), t) ={}& \partial_x \left[\log\rho_\ast(x(t), t)\right]\cdot \frac{\dd x(t)}{\dd t} +   \partial_t \log\rho_\ast(x(t), t) \nonumber \\
    ={}& S_\ast(x(t), t) \cdot v_\theta(x(t),t) + \partial_t {\delta} \log \tilde{\rho}_\ast(x(t),t),
\end{align}
where we have used $\rho_\ast(x,t) = \tilde{\rho}_\ast(x,t)/\mathcal{Z}(t)$ and Eq.~\eqref{eq:source}. From the above equations, it is clear that the log-densities along the trajectories can be computed along the dynamics via
\begin{align}
    \log \rho_\theta(x(t), t) ={}& \log \rho_\theta(x(0),0) - \int_0^t \nabla \cdot v_\theta\left(x(s),s\right)\, \dd s,  \label{eq:logrhotheta} \\
    \log \rho_\ast(x(t), t) ={}& \log \rho_\ast(x(0),0) + \int_0^t  \left[S_\ast(x(s), s) \cdot v_\theta(x(s),s) + \partial_t {\delta} \log \tilde{\rho}_\ast(x(s),s) \right]\, \dd s \label{eq:logrhoast} 
\end{align}
We remark that Eq.~\eqref{eq:logrhotheta} is the pivotal construct to enable density estimation by Neural ODE \citep{neuralODE}. Because the initial distribution $\rho_\theta(\cdot, 0) = \mu(0) = \rho_\ast(\cdot, 0)$ we get the following important result,

With $\rho_\theta$ and $\rho_\ast$ defined as above, we have the following density ratio
\begin{align}
    \log \frac{\rho_\ast(x(t), t) }{\rho_\theta(x(t), t)} = {}&  \int_0^t \left[ \nabla \cdot v_\theta\left(x(s),s\right) + S_\ast(x(s), s) \cdot v_\theta(x(s),s) + \partial_t {\delta} \log \tilde{\rho}_\ast(x(s),s) \right]\,  \dd s. \label{eq:densityRatio}
\end{align}
The above equation hints that we should define a dynamic error function along a specific sample trajectory
\begin{equation}
    \varepsilon(t;x_0) \equiv \nabla \cdot v_\theta\left(x(t;x_0),t\right) + S_\ast(x(t;x_0), t) \cdot v_\theta(x(t;x_0),t) + \partial_t {\delta} \log \tilde{\rho}_\ast\left(x(t;x_0), t \right), \label{eq:varepsilon}
\end{equation}
which is identically the difference between the LHS and RHS of Eq.~\eqref{eq:consistent} along the trajectory. We remark that the error function defined in \eqref{eq:training} is consistent to the definition \eqref{eq:varepsilon} above, but we have dropped the time-dependence in Eq.~\eqref{eq:training}. Then, for every trajectory $x(t)$, we associate a dynamic weight $w(t;x_0)$ to the trajectory
\begin{equation}
    \tilde{w}(t;x_0) := \frac{\rho_\ast(x(t;x_0), t) }{\rho_\theta(x(t;x_0), t)} = \exp\left(\int_0^t \varepsilon(s;x_0)\, \dd s.\right) \label{eq:weights}
\end{equation}
Note that because $x(t;x_0)$ depends on the initial condition $x_0$, the dynamic error function $\varepsilon(t;x_0)$ and the difference between the log-densities also depend on the initially sampled $x_0 \sim \mu$. 

\begin{proposition}\label{prop:optimality}
    Assuming $\rho_\theta$ dominates $\rho_\ast$. In the optimal case when $\varepsilon(x,t)=0$ $\forall x$ in the support of $\rho_\theta$, the density function $\rho_\theta$ induced  by the (trained) neural-network-modeled flow is identical to that of the target distribution, $\rho_\ast$. In this case, $\tilde{w}=1$.
\end{proposition}
\begin{proof}
    By Eqs.~\eqref{eq:densityRatio} and \eqref{eq:weights}. 
\end{proof}

Equation \eqref{eq:weights} plays a key role in the following analysis. 

\begin{theorem}\label{lem:liouville-mean}
Given a neural velocity field $v_\theta(x,t)$ and an integrable function $F(x)$. Then for any $t\ge 0$,
\begin{equation}
    \mathbb{E}_{x \sim \rho_\theta\left(\cdot, t\right)} \left[ F(x) \right] = \mathbb{E}_{x_0  \sim \mu } \left[ F\left(x\left(t;x_0\right)\right) \right], 
\end{equation}
where $\rho_\theta\left(\cdot, t\right)$ is the solution of the following Generalized Liouville Equation 
\begin{equation}
    \partial_t \rho_\theta\left(x,t\right)= - \nabla_x \cdot \left[ v_\theta\left(x,t\right) \rho_\theta\left(x,t\right) \right]
\end{equation}
with initial data
\begin{equation} 
    \rho_\theta\left(x,0\right) = \mu(x). 
\end{equation}
\end{theorem}
\begin{proof}
This Lemma follows directly from the definition of $\rho_\theta(\cdot, t)$ in the Generalized Liouville Equation, that $\rho\left(x,t\right)$ is the density of the trajectory $x(t)$ with an initial condition $x_0 \sim \mu$. See \citet{gerlich1973VerallgemeinerteLiouvilleGleichung}. 
\end{proof}
\begin{remark} 
    Lemma \ref{lem:liouville-mean} is the theoretical foundation of Neural ODE \cite{neuralODE}, a continuous-time normalizing flow.
\end{remark}

Next, we show that the ratio of the log-densities in Eq.~\eqref{eq:weights} provides a way for us to estimate the expectation of a measurable function $F$ with respect to the target density $\rho_\ast\left(\cdot, t\right)$ by a change-of-measure, without the need of generating samples from the target distribution. 

\begin{lemma}\label{lem:change_of_measure}
Assuming $\rho_\theta$ dominates $\rho_\ast$, for any $t\in \left[0,1\right]$,
\begin{align}
    \mathbb{E}_{x'\sim \rho_\ast (\cdot, t)} \left[F(x')\right] = \mathbb{E}_{x_0\sim \mu } \left[\tilde{w}(t;x_0)\, F\left(x\left(t;x_0\right)\right) \right]. 
\end{align}
\end{lemma}
\begin{proof}
    \begin{align}
    \mathbb{E}_{x'\sim \rho_\ast (\cdot, t)} \left[F\left(x'\right)\right] ={}&  \int F\left(x'\right) \rho_\ast(x',t)\, \dd x' \nonumber \\
    ={}&   \int F\left(x'\right) \frac{\rho_\ast(x',t)}{\rho_\theta(x',t)}\rho_\theta(x',t)\,  \dd x' \nonumber \\
    ={}& \mathbb{E}_{x'\sim \rho_\theta (\cdot, t)} \left[F\left(x'\right) \frac{\rho_\ast(x',t)}{\rho_\theta(x',t)}\right] \nonumber \\
    ={}& \mathbb{E}_{x_0\sim \mu } \left[F\left(x\left(t;x_0\right)\right) \frac{\rho_\ast(x\left(t;x_0\right),t)}{\rho_\theta(x\left(t;x_0\right),t)}\right] \nonumber \\
    ={}& \mathbb{E}_{x_0\sim \mu } \left[ \tilde{w}\left(t;x_0\right)\, F\left(x\left(t;x_0\right)\right) \right],
\end{align}
where we used Lemma \ref{lem:liouville-mean} to establish the second-to-last equality and the definition Eq.~\eqref{eq:weights} to establish the last equality. 
\end{proof}

\begin{remark} 
    We put a stringent condition that $\rho_\theta$ has to dominate $\rho_\ast$ for any arbitrary function $F$. The condition is often relaxed to ``$\rho_\theta(x) > 0$ where $F(x) \rho_\ast(x) > 0 $'' in the importance sampling literature (e.g., \citet{tokdar2010importance}), which requires knowledge of the function $F$. 
\end{remark}

\begin{lemma}\label{lem:source}
Assuming $\rho_\theta$ dominates $\rho_\ast$ and $\partial_t \log \tilde{\rho}_\ast(x,t)$ is integrable with respect to $\rho_\ast(\cdot, t)$ for any $t\in \left[0,1\right]$,
\begin{align}
    \mathbb{E}_{x'\sim \rho_\ast (\cdot, t)} \left[\partial_t \log \tilde{\rho}_\ast(x',t) \right] = \mathbb{E}_{x_0\sim \mu } \left[\tilde{w}(t;x_0)\, \, \partial_t \log  \tilde{\rho}_\ast\left(x(t;x_0),t\right) \right].
\end{align}
\end{lemma}
\begin{proof}
    By Lemma \ref{lem:change_of_measure} with $F(x):=\partial_t \log \tilde{\rho}_\ast(x,t)$. 
\end{proof}

\begin{proposition}\label{prop:estimator}
    Given an integrable function $F$, $\mathbb{E}_{x'\sim \rho_\ast (\cdot, t)} \left[F\left(x'\right) \right] $ can be unbiasedly estimated by a finite set of initial samples $x_0^{(i)}$, 
    \begin{equation}
        \mathbb{E}_{x'\sim \rho_\ast (\cdot, t)} \left[F\left(x'\right) \right] \overset{\text{i.p.}}{\longrightarrow} F_N(t):=\sum_{i=1}^{N} w_i(t) \, F\left(x\left(t;x_0^{(i)}\right)\right),
    \end{equation}
    where normalized weights $w_i(t)$ are defined as (Eq.~\eqref{eq:weights-main}):
    \begin{equation}
    {w}_i(t) := \frac{ \tilde{w}\left(t, x_0^{(i)}\right)}{ \sum_{j=1}^{N} \tilde{w}\left(t, x_0^{(j)}\right)}. 
\end{equation}

\begin{remark}
Note that the convergence result using the weight normalization (Eq.~\eqref{eq:weights-main}) rests on the following fact
    \begin{equation*}
   \lim_{N\to \infty}\frac{1}{N}\sum\limits_{i=1}^{N}\tilde{w}(t;x_0^{(i)})F(x(t;x_0^{(i)})) = \lim_{N\to \infty}\sum_{i=1}^{N} w_i(t) \, F\left(x\left(t;x_0^{(i)}\right)\right)
\end{equation*}
which follows from importance sampling. See~\cite{geweke1989bayesian} for the proof and~\cite{tokdar2010importance} for a review.
\end{remark}

\begin{remark}\label{rmk:strong_conv_weights}
    Note that in practice we get the stronger almost sure convergence in the result above by demanding that $\mathbb{E}_{x_0\sim \mu } \lvert \,\tilde{w}\left(t;x_0\right)\, F\left(x\left(t;x_0\right)\right) \, \rvert < \infty$. This is not an unreasonable assumption as each weight $\tilde{w}$ is absolutely bounded when using a suitably expressive neural network.
\end{remark}

\end{proposition}

\begin{corollary}\label{cor:estimatingPartialtLog}
 Assuming $\partial_t \log\tilde{\rho}_\ast \left(x,t\right)$ is integrable, $\mathbb{E}_{x'\sim \rho_\ast (\cdot, t)} \left[ \partial_t \log\tilde{\rho}_\ast \left(x,t\right) \right] $ can be unbiasedly {asymptotically (i.e.~unbiased as $N\rightarrow \infty$)} estimated by a finite set of initial samples $x_0^{(i)}$, 
    \begin{equation}
        \mathbb{E}_{x'\sim \rho_\ast (\cdot, t)} \left[F\left(x'\right) \right] \approx \sum_{i=1}^{N} w_i(t) \, \partial_t \log\tilde{\rho}_\ast \left(x\left(t;x_0^{(i)}, t\right)\right).
    \end{equation}
\end{corollary}

\begin{corollary}
The last term in Eq.~\eqref{eq:varepsilon}, $\partial_t {\delta} \log \tilde{\rho}_\ast \left(x(t), t\right) $, can be estimated at time $t$. 
\end{corollary}
\begin{proof}
By Eq.~\eqref{eq:source}, $\partial_t {\delta} \log \tilde{\rho}_\ast \left(x(t), t\right)$ contains two terms: $\log \tilde{\rho}_\ast \left(x^{(i)}\left(t\right),t\right)$, and $\mathbb{E}_{x'\sim \rho_\ast (\cdot, t)} \left[\partial_t \log \tilde{\rho}_\ast(x',t) \right]$. The former can be evaluated straightforwardly as the density function is given. The latter can be estimated by Corollary \ref{cor:estimatingPartialtLog}.
\end{proof}

\section{(Asymptotically) Unbiased Estimators of the Marginalized Likelihood} \label{app:logZ}

\begin{proof}{Theorem \ref{thm:logZ}}

First, we show that that $\log \mathcal{Z}(t)$ is given by integrating $I(s)$ on $[0,t]$ where $I(s) = \mathbb{E}_{\rho_\ast\left(\cdot, s\right)} \left[\partial_{s} \log \tilde{\rho}_\ast \left(x,s\right) \right]$ i.e.
\begin{equation}
    \log \mathcal{Z}(t) = \int_{0}^t \mathbb{E}_{\rho_\ast\left(\cdot, s\right)} \left[\partial_{s} \log \tilde{\rho}_\ast \left(x,s\right) \right]\, \dd s.
\end{equation}
This can be established by first applying $\dd /\dd t$ to $\log\mathcal{Z}\left(t\right):=\int \tilde{\rho}_\ast\left(x,t\right) \dd x$:
\begin{align}\label{eq:logZ_int}
    \frac{\dd}{\dd t } \log \mathcal{Z}(t) ={}& \frac{\dd}{\dd t} \log \int \tilde{\rho}_\ast \left(x, t\right)\, \dd x = \frac{1}{\mathcal{Z}(t)} \int \frac{1}{\tilde{\rho}_\ast \left(x, t\right) } \left(\frac{\partial \tilde{\rho}_\ast \left(x, t\right) }{\partial t}  \right) \tilde{\rho}_\ast \left(x, t\right) \, \dd x  \nonumber \\
    ={}& \int \frac{\partial  \log\tilde{\rho}_\ast \left(x, t\right) }{\partial t} \frac{\tilde{\rho}_\ast \left(x, t\right)}{\mathcal{Z}(t)} \, \dd x
    =\mathbb{E}_{x\sim \rho_\ast\left(\cdot, t\right)} \left[\partial_{t} \log \tilde{\rho}_\ast \left(x,t\right) \right].
\end{align}
Integrating the above equation from $0$ to $t$, we obtained
\begin{equation}
    \log \mathcal{Z}(t) = \int_{0}^t \mathbb{E}_{\rho_\ast\left(\cdot, s\right)} \left[\partial_{s} \log \tilde{\rho}_\ast \left(x,s\right) \right]\, \dd s, \label{eq:pathIntegrallogZ}
\end{equation}
because $\log \mathcal{Z}(0)=0$ a we fix $\rho_\ast(t=0)=\mu$, a normalized distribution. We remark that this derivation is akin to that in \cite{gelmanSimulatingNormalizingConstants1998}. 
Corollary \ref{cor:estimatingPartialtLog} provides us with an unbiased estimate of $I(s)$ using normalized weights as in Eq.~\eqref{eq:weights-main}. Thus we have,
\begin{equation}
    {\lim_{N\rightarrow \infty}} \sum_{i=1}^N w_i\left(t\right) \, \partial_t\log \tilde{\rho}\left(x^{(i)}\left(t\right), t\right) \overset{\text{i.p.}}{\longrightarrow} I(s),
\end{equation}
for each $s\in [0,t]$. In practice as remark~\ref{rmk:strong_conv_weights} suggests, we do get stronger almost sure convergence. However, given the assumption that both the dynamic weights $w_i(t)$ and $\partial_t \log \tilde{\rho}_{\ast}\left(x^{(i)}(t), t\right)$ are absolutely bounded almost everywhere along the sample trajectories our final result still holds. 
With this assumption in mind, we observe that the convergence of the integrand in probability 
\begin{equation}
    {\lim_{N\rightarrow \infty}} \sum_{i=1}^N w_i\left(t\right) \, \partial_t \log \tilde{\rho}\left(x^{(i)}\left(t\right), t\right) \overset{\text{i.p.}}{\longrightarrow} \mathbb{E}_{\rho_\ast\left(\cdot, s\right)} \left[\partial_{s} \log \tilde{\rho}_\ast \left(\cdot,s\right) \right]
\end{equation}
implies the almost-surely convergence of the integration via the Lebesgue Dominated Convergence Theorem (see Corollary 6.3.2 in~\cite{resnick2019probability}). Thus, we have
\begin{equation}
     {\lim_{N\rightarrow \infty}}  \int_{0}^{t} \sum_{i=1}^N w_i\left(s\right) \, \partial_s \log \tilde{\rho}\left(x^{(i)}\left(s\right), s\right) ds \overset{\text{a.s.}}{\longrightarrow} \int_{0}^{t}\mathbb{E}_{\rho_\ast\left(\cdot, s\right)} \left[\partial_{s} \log \tilde{\rho}_\ast \left(\cdot,s\right) \right] \dd s \overset{\text{Eq.~\eqref{eq:pathIntegrallogZ}}}{=} \log \mathcal{Z}\left(t\right). 
\end{equation}

\end{proof}

\begin{remark}
    The boundedness assumption in Theorem~\ref{thm:logZ} can be relaxed to require a dominating integrable function. However, in practice we do see the assumptions met especially if we have a suitably expressive neural velocity model.
\end{remark}

{
We now show the proof of the unbiased estimation of $\mathcal{Z}$, Theorem \ref{thm:Z}.

\begin{proof}{Unbiased estimator of $\mathcal{Z}$.}
We first relate $\varepsilon$ defined in \eqref{eq:staticError} and $\epsilon$ in \eqref{eq:modifiedError}:
\begin{equation}
\epsilon\left(x\left(s\right),\theta\right) = \varepsilon\left(x\left(s\right),\theta\right) + \partial_s \left\langle  \log \tilde{\rho}_\ast(\cdot,s) \right \rangle_\ast,
\end{equation}
because of the definition Eq.~\eqref{eq:source}. Integrating both sides from $s=0\rightarrow t$ and using Eqs.~\eqref{eq:modifiedLogWeights} \eqref{eq:varepsilon}, \eqref{eq:weights}, and \eqref{eq:pathIntegrallogZ}, we obtain
\begin{equation}
    \log \varpi(t;x_0) = \log {w}(t;x_0) + \log \mathcal{Z}(t). 
\end{equation}
Exponentiating the above expression leads to
\begin{equation}
    \varpi(t;x_0) = {w}(t;x_0)\, \mathcal{Z}(t).
\end{equation}
Now, we perform expectation over the empirical distribution $\rho_\theta$ and using Eq.~\eqref{eq:weights}:
\begin{equation}
    \mathbb{E}_{\rho_\theta} \left[ \varpi(t;x_0) \right] =  \mathbb{E}_{\rho_\theta} \left[ {w}(t;x_0)\, \mathcal{Z}(t) \right] = \mathcal{Z}(t) \,\mathbb{E}_{\rho_\theta} \left[ {w}(t;x_0) \right] = \mathcal{Z}(t) \, \mathbb{E}_{\rho_\ast} \left[ t \right] = \mathcal{Z}(t). 
\end{equation}
We can now unbiasedly estimate $\mathcal{Z}(t)$ using Monte Carlo, i.e., Eq.~\eqref{eq:Zestimate}. 
\end{proof}

}

\begin{remark}
    Our results are in continuous time and the Generalized Liouville Equation requires that we have the exact velocity field. In practice, we use a time integrator to evolve the velocity field in discrete time in chunks of $\delta t$. In this paper we use (Forward) Euler time stepping. Hence the results above do not hold in general; in fact there is a bias term which depends on $\delta t$. However, in the limit as $\delta t \to 0$ we do get the unbiased estimate (assuming that our neural network modeled velocity is Lipschitz and not chaotically dependent on the initial condition). 
\end{remark}

\section{Additional Experiment Details and Results} \label{app:additionalDetails}
In this section, we provide additional details of the numerical experiments for reproducibility. 

\subsection{Additional details of the test distributions}

For the Mode-separated Gaussian mixture problems, we followed the testing problem proposed in \citet{zhang2021path} and chose a small variance ($\sigma=0.012$) for each mode of the Gaussian mixture to make it more challenging for different methods. We reduce the grid size to $\{ -1,0,1\}^2$ such that using the simplest prior/reference distribution $\mathcal{N}(0,\bf{I})$ for different methods could provide good coverage of the target distribution. This testing problem could provide a fair comparison between different methods for samples from well-separated modes in a distribution. Our results show that the proposed LFIS excels at generating evenly distributed samples from different modes of the target distributions. 

For the Log Gaussian Cox process, the covariance matrix of the prior distribution $K$ takes the form $K(u,v) = \sigma^2 \exp\l ( -\frac{\vert u-v\vert_2}{M \beta}\r)$, where $\sigma^2 = 1.91$, $M=40$ is the grid number in each direction, and $\beta=1/33$. Here, $u$ and $v$ denote the normalized positions on two grid points. The mean vector of the prior distribution is $\log(126) - \sigma^2$.

For the sampling problem in the latent space of VAE, we took the decoder $p_\theta(x\vert z)$ of the pre-trained VAE in \citet{AFT} and used it to construct the likelihood function with binary cross-entropy loss. In Fig.~\ref{fig:VAE}, we demonstrate the test image and the reconstructed image using the pre-trained decoder $p_\theta(x\vert z)$ from samples generated by LFIS.

\begin{figure}[htb] 
    \centering
    \includegraphics[width=0.80\textwidth]{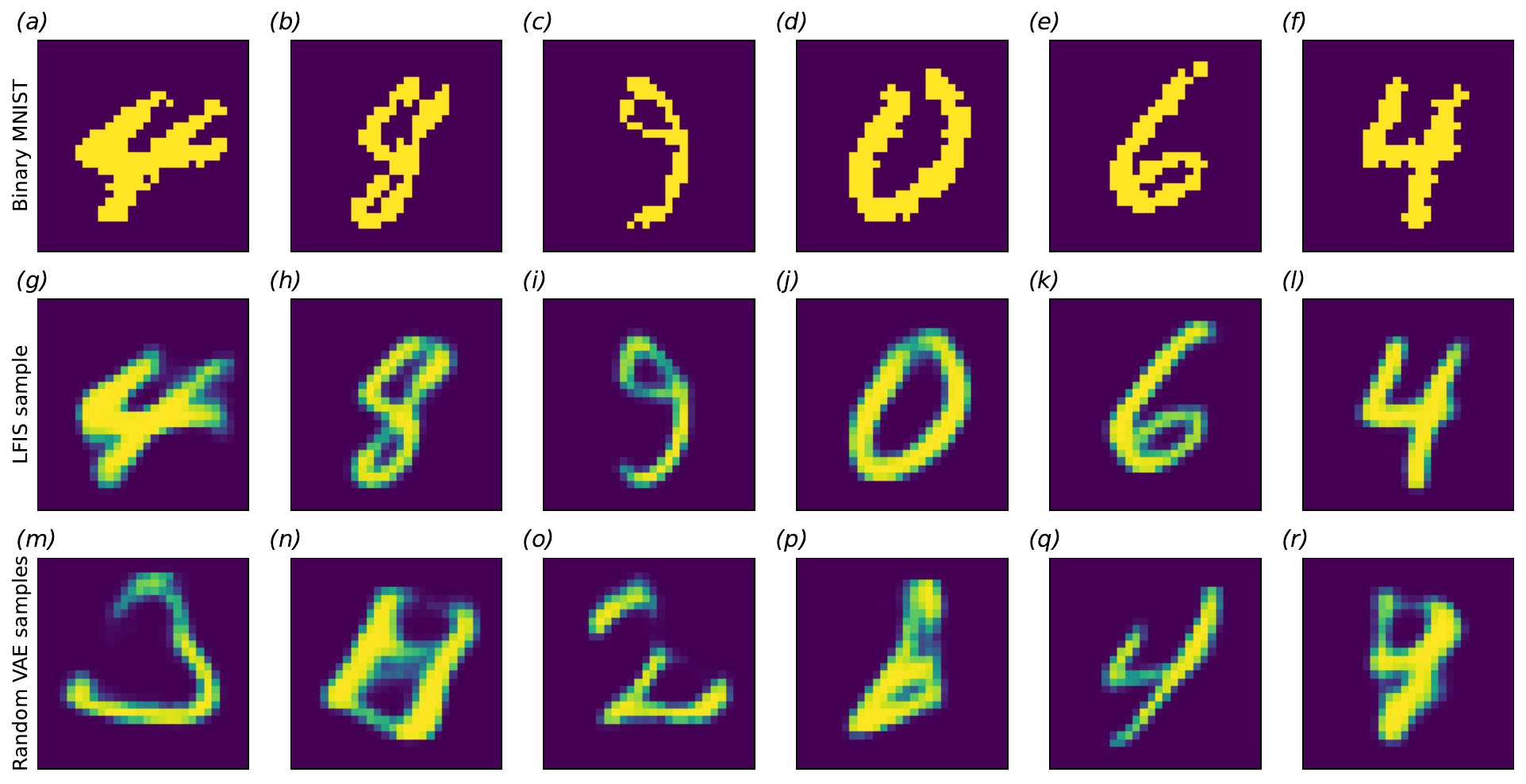}
    \caption{Binary MNIST dataset (\textit{a-f}), decoded LFIS samples (\textit{g-l}), and decoded image from random latent space samples (\textit{m-r}).}
    \label{fig:VAE}
\end{figure}

{
\subsection{LFIS for multi-modal distributions with unequally weighted modes} \label{app:MG2DWeights}

In this section, we test LFIS on distributions with unequally weighted modes. We reformulate the mode-separated Gaussian mixture distribution and give each mode different weights. We tested three different configurations of unequally weighted modes whose PDFs are shown in Figs.~\ref{fig:MG2DWeights} (a,d,g). The numerical experiments are performed using the exact setup as the regular Gaussian mixture distribution. The samples drawn from the ground-truth distributions and LFIS are shown Fig.~\ref{fig:MG2DWeights} (b,c,e,f,h,i). The $\log \mathcal{Z}$ estimation for each configuration is shown in table \ref{tab:MG2DWeights}. Evidently, LFIS is capable of sampling from multi-modal distributions with unequally weighted modes.

\begin{figure}[htb] 
    \centering
    \includegraphics[width=0.80\textwidth]{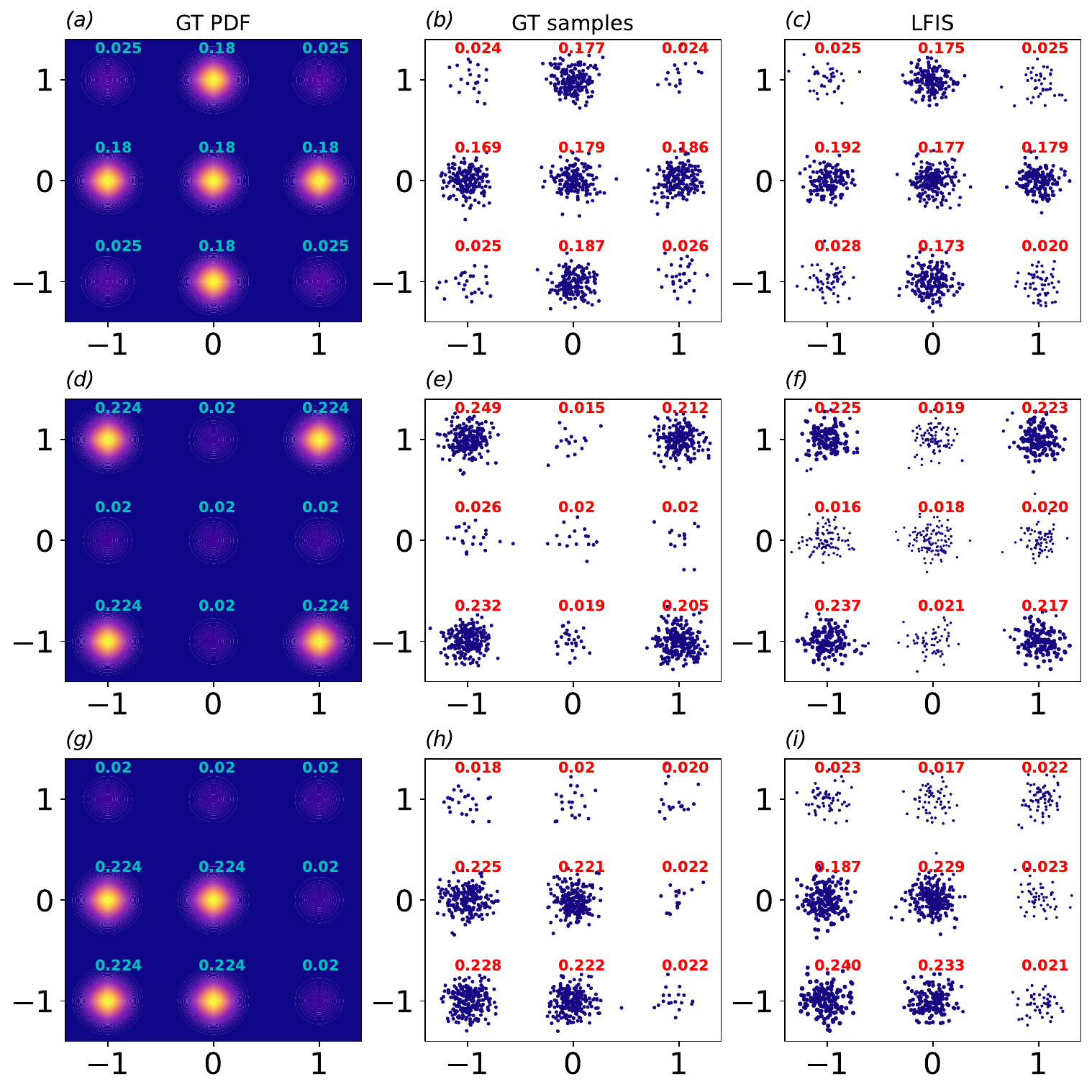}
    \caption{{Mode-separated Gaussian mixture with mixed weights for each mode. Three different configures of the mixed weights are considered. Subfigures (\textit{a,d,g}) show the ground-truth PDFs with the corresponding weights annotated by each mode. The samples generated from the ground-truth PDF and LFIS are shown in subfigures (\textit{b,e,h}) and (\textit{c,f,i}), with the sizes of the samples scaled by the sample weights (note that the sample weights are different from the modal wights).}} 
    \label{fig:MG2DWeights}
\end{figure}

\begin{table}[!t]
\caption{{$\log \hat{\mathcal{Z}}$ estimation for mode-separated Gaussian mixture with unequal weights.}}
\vskip 0.15in
\centering
\begin{small}
\begin{tabular}{c c c}
\toprule
  & MG ($D=2$)  \\ 
\midrule 
Configuration 1 & -0.0016 $\pm$ 0.005  \\
Configuration 2 & 0.0013 $\pm$ 0.004  \\
Configuration 3 & -0.0003 $\pm$ 0.006  \\
\bottomrule
\end{tabular}
\label{tab:MG2DWeights}
\end{small}
\end{table}

}

\subsection{Neural network architecture}

The expressibility of the neural network could significantly impact the sampling quality of the LFIS, considering that LFIS imposes a specific density flow from $t=0 \rightarrow 1$ to satisfy $\rho_\ast(x,t)$, while other methods like PIS, DDS do not enforce this condition. However, in the numerical experiments, we find that an NN similar to those used in PIS and DDS is expressive enough for all the testing problems. For LFIS, we use an NN with 2 hidden layers, each with 64 nodes to parameterize the discrete-time velocity field. For both DDS and PIS, the NNs have time-encoding and are augmented by the gradient information or the score $S(x,t)$. 

It is difficult to make a direct comparison of NN structures and numbers of parameters between the discrete-time and continuous-time models. For LFIS, the total number of parameters will increase with the total number of steps. For DDS and PIS, the number of parameters will increase with the number of channels used for time-encoding. {Overall, when LFIS uses the same number of time steps as the time-encoding in DDS and PIS, the total number of NN parameters are very similar, thus causing no additional memory footprint.}

\subsection{Additional details of the training procedure}

The general training procedure is summarized in Algorithm \ref{alg:1}. For each discretized time step $k$, at each training epoch, a new minibatch (B) of samples is generated by calling the \texttt{GenerateSamples} function. This step is crucial to provide good coverage of the sampling space. However, for low-dimensional problems, repeatedly calling the \texttt{GenerateSamples} function for each minibatch could slow down the training process, because the initial samples need to be passed through all the previously trained flow $v_{\theta_\ast^{(l)}}$. To accelerate the training of low-dimensional sampling problems ($D<10$), we skip the generation of new samples at each epoch. Instead, we use a subset of the $N$ samples used for estimating $\langle \partial_t \log \tilde{\rho_\ast}\rangle_\ast$ for training. When $N$ is large enough to provide good coverage of the sampling space, this simplified training procedure could provide a noticeable acceleration of learning the Liouville flow. 

The choice of stopping criteria for training the Liouville flow at each time step $v_{\theta_\ast^{(l)}}$ is also crucial for improving the training quality and speed. Through a series of numerical experiments, we introduce the criteria $\mathbb{E}_i\left [ \varepsilon^2(x^{(i)}(t);\theta) \right ]/\text{var}_i\left[\partial_t \log\tilde{\rho}_\ast \left(x^{(i)}(t), t\right)\right]$ at each training time, which represents the percentage of the squared error to the total variation of the RHS of the equation. In general, we found that when the criteria converge to less than 0.001 during training, the learned velocity field can provide a good approximation of the Liouville flow. If the proposed criteria can not be met during training, either due to the limited expressibility of the NN, or accumulated error from the previous velocity field, we will stop the training at 2,000 epochs. Empirically, we found the NNs converged to the desired 0.001 for at least half of the discrete timesteps, for most of the test problems.  The imperfect representation of the Liouville flow using NN will be quantified using the sample weights during sampling.

For all the numerical experiments, we use the Adam optimizer with an initial learning rate of $5\times 10^{-3}$. We employ an optimizer schedule that will reduce the learning rate to 50\% every 200 epochs without observing any improvement in the loss.

\subsection{Choice of schedule} \label{app:scheduling}

\begin{figure}[htb] 
    \centering
    \includegraphics[width=0.89\textwidth]{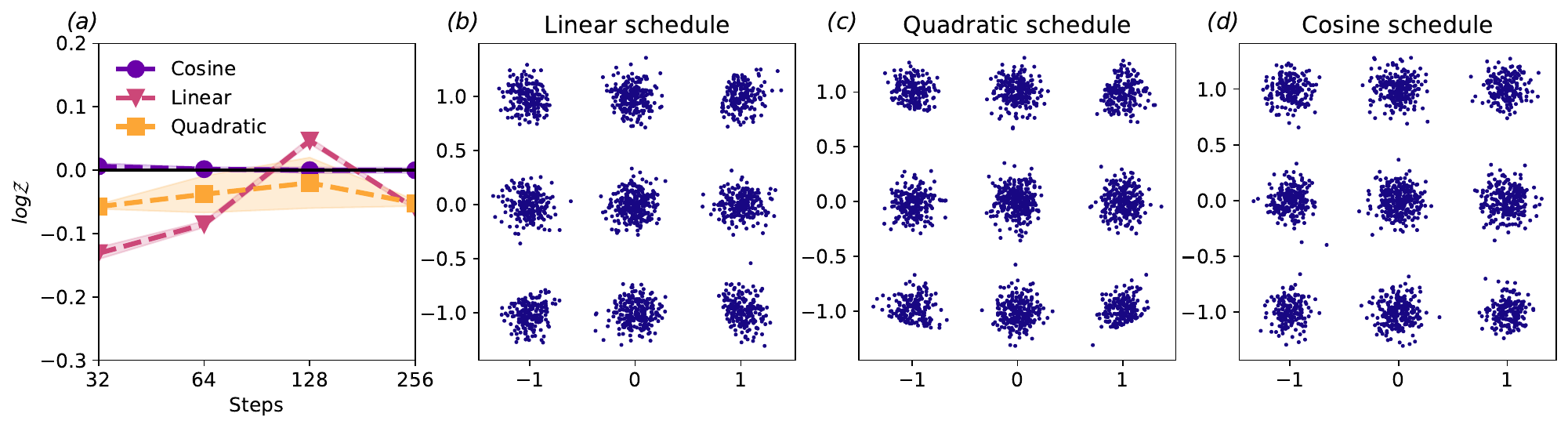}
    \caption{The effects of schedule on $\log\hat{\mathcal{Z}}$ estimation at different $T$s and sample quality.}
    \label{fig:schedule}
\end{figure}

We tested three different choices of the schedule function $\tau(t)$, which controls the ``pace'' of the annealing: (1) a linear schedule $\tau(t)=t$ (with an even ``pace''), (2) a quadratic schedule $\tau(t)=t$ (a slow pace at the beginning when $\rho_\ast(t)$ is close to $\mu$ and fast in the end when  $\rho_\ast(t)$ is close to $\nu$), and (3) a cosine schedule $\tau(t)=\left(1-\cos(\pi t)\right)/2$ (slow pace at the beginning when $\rho_\ast(t)$ is close to $\mu$ and in the end when  $\rho_\ast(t)$ is close to $\nu$ and fast during intermediate times.) The numerical experiment results are shown in Table \ref{tab:schedule} and Fig.~\ref{fig:schedule} following the same procedure as described above for estimating $\log\mathcal{Z}$ for the 2D Gaussian mixture problems. As we can see in Fig.~\ref{fig:schedule} (a), the cosine schedule shows better convergence of $\log\mathcal{Z}$ as we increase the number of total time steps. In Fig.~\ref{fig:schedule} (b-d), we show the samples generated using three types of schedules. We observe that the samples from both the linear schedule and quadratic schedule show non-uniform coverage at the edge of the distribution, while the samples generated with the cosine schedule uniformly span over each mode of the Gaussian mixture distribution. Based on these tests, we use the cosine schedule for the rest of the numerical experiments for LFIS.

\begin{table}[!t]
\caption{The effects of schedule on $\log \hat{\mathcal{Z}}$ estimation.}
\vskip 0.15in
\centering
\begin{small}
\begin{tabular}{c c c}
\toprule
 Schedule  & MG ($D=2$)  \\ 
\midrule 
Linear & -0.06 $\pm$ 0.003  \\
Quadratic & -0.05 $\pm$ 0.005  \\
Cosine & -0.0002 $\pm$ 0.004  \\
\bottomrule
\end{tabular}
\label{tab:schedule}
\end{small}
\end{table}

\subsection{Improvement of statistical estimation by sample weights}
\label{app:correction}
A novel contribution of LFIS is using the accumulated error induced by imperfectly trained NN as sample weights for unbiased and consistent estimation of statistical quantities. In this section, we provide numerical results showing the improvement of statistics estimation using accumulated errors as weights, using the MG and Funnel test problems which are analytical. In the proposed LFIS algorithm, the weights are used in both training and sampling procedures for computing $\langle \partial_t \log \tilde{\rho_\ast}\rangle_\ast$. To measure the effects of sample weights on training and sampling separately, we performed four different combinations of numerical experiments for both MG and funnel distributions. We compare the $\log \mathcal{Z}$ estimation in Table \ref{tab:weights}. There are no significant improvements in adding sample weights during training. However,
by using the accumulated errors as sample weights during sampling, the estimations can be significantly improved in these two models. The results evidentiate the effectiveness of our proposed novel corrections for sampling.

\begin{table}[!t]
\caption{$\log \hat{\mathcal{Z}}$ estimation of LFIS with/without weights}
\vskip 0.15in
\centering
\begin{small}
\begin{tabular}{c c c c c c}
\toprule
  & & \multicolumn{2}{c}{Training w/ weights}  & \multicolumn{2}{c}{Training w/o weights}\\ 
&$T$ & Sampling w/ weights & Sampling w/o weights & Sampling w/ weights &Sampling w/o weights \\ 
\midrule
&32  & $0.0060 \pm 0.005$ & $-0.130 \pm 0.050$ & $0.0048 \pm 0.005$ & $-0.150 \pm 0.061$ \\
MG &64 & $0.0015 \pm 0.003$  & $-0.076 \pm 0.051$ & $0.0009 \pm 0.004$  & $-0.058 \pm 0.045$ \\
($D=2$)&128  & $-0.0001 \pm 0.004$ & $-0.032 \pm 0.066$ & $0.0001 \pm 0.004$ & $-0.042\pm 0.064$ \\
&256  & $-0.0002 \pm 0.004$ & $-0.035 \pm 0.070$& $0.0002 \pm 0.004$ & $-0.028 \pm 0.044$\\
\midrule
&32  & $-0.29 \pm 0.022$ & $-0.45 \pm 0.052$ & $-0.56 \pm 0.002$ & $-0.61 \pm 0.044$\\
Funnel & 64 & $-0.159 \pm 0.028$ & $-0.31 \pm 0.053$ & $-0.16 \pm 0.009$ & $-0.29 \pm 0.043$\\
(D=10)&128  & $-0.19 \pm 0.01$ & $-0.27 \pm 0.070$& $-0.15 \pm 0.010$ & $-0.24 \pm 0.048$\\
&256  & $-0.07 \pm 0.003$ & $-0.15 \pm 0.045$& $-0.06 \pm 0.018$ & $-0.15 \pm 0.065$\\
\bottomrule
\end{tabular}
\label{tab:weights}
\end{small}
\end{table}

\subsection{Training and sampling cost}

Similar to PIS and DDS, the proposed LFIS is an NN-parameterized model that can be trained once and then deployed for sampling, while the MC-based algorithms need to be repeated every time they are used to perform sampling. In this section, we focus on comparing training costs and sampling costs within similar type of algorithms that are based on NN. 

For LFIS, DDS, and PIS, we performed 30 independent training and sampling processes for the 10D funnel distributions. We set the total steps to be $T=256$ and trained the NNs of different methods for a maximum of 2,000 epochs or until the convergence criteria were met for each of the methods. The batch size is chosen to be 10,000. The optimizer, learning rate, and learning rate schedule are set to be the same as provided by the original papers. For the sampling experiments, we took the pre-trained models and deployed them to independently generate 30 batches of samples, each with 2,000 points. All the experiments are performed using a single NVIDIA A100 GPU with 40GB of RAM.

Table \ref{tab:cost} compares the training and sampling time of different NN-based sampling methods. For training, the LFIS is faster than both PIS and DDS, which is partially due to the proposed stop criteria for the convergence of NN at each discrete time step. On the other hand, DDS has the fastest sampling time because of the probability flow ODE and the just-in-time (JIT) compilation with JAX. However, the first run/compilation of the DDS took 4942ms (with weight computation) and 1528ms (w/o weight computation). As such, DDS with JIT has a worse performance unless one desires multiple sampling with the same trained model. For LFIS, the majority of the sampling time is attributed to the computation of sample weights. In our implementation of LFIS based on PyTorch, the divergence of the velocity field is computed via the \texttt{torch.autograd.functional.jacobian} function, which is not optimized for divergence computation, nor compilable currently.

\begin{table}[!t]
\caption{Traning and sampling cost}
\vskip 0.15in
\centering
\begin{small}
\begin{tabular}{c c c}
\toprule
 Method  & Training time (minutes) & Sampling time (ms)  \\ 
\midrule 
LFIS (w/o weight, not compiled)  & $27.6 \pm 1.84$ & $90.4 \pm 4.18$ \\
LFIS (w/ weight, not compiled)  & - & $526.9 \pm 7.48$  \\
PIS & $36.4 \pm 1.51$  & $819.9 \pm 3.99$ \\
DDS (w/o weight, JIT)  & $37.13 \pm 0.28$ & $51.91 \pm 2.27$ \\
DDS (w/ weight, JIT)  & - & $57.10 \pm 6.09$ \\
\bottomrule
\end{tabular}
\label{tab:cost}
\end{small}
\end{table}

\subsection{Reproducibility of LFIS training}

We demonstrate the reproducibility of the LFIS sampling statistics by performing the 30 independent trainings from different initializations of the NN parameters. For each of the independently trained LFIS, we run 30 sampling procedures of 2,000 samples to estimate the mean and variance of $\log\mathcal{Z}$. In Fig.~\ref{fig:repeatTraining}, we plot the histograms of $\log\mathcal{Z}$ statistics for 30 LFISs. We observe a very narrow spread of $\log\mathcal{Z}$ statistics over all 30 trainings, which shows the reproducibility of the proposed methods. In the main text, we report the statistics that is the closest to the mean of 30 LFISs.

\begin{figure}[htb] 
    \centering
    \includegraphics[width=0.65\textwidth]{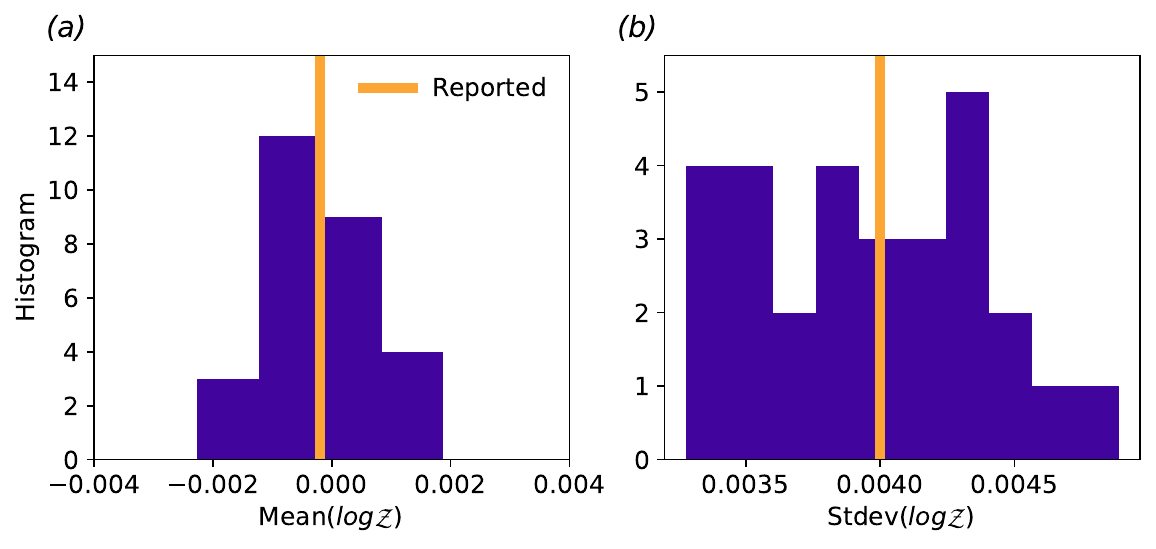}
    \caption{$\log\hat{\mathcal{Z}}$ estimates from 30 independent Liouville flow training for MG (D=2) problem.}
    \label{fig:repeatTraining}
\end{figure}

\subsection{Choice of total time steps and tempered scales} \label{app:Tsteps}

\begin{figure}[htb] 
    \centering
    \includegraphics[width=0.75\textwidth]{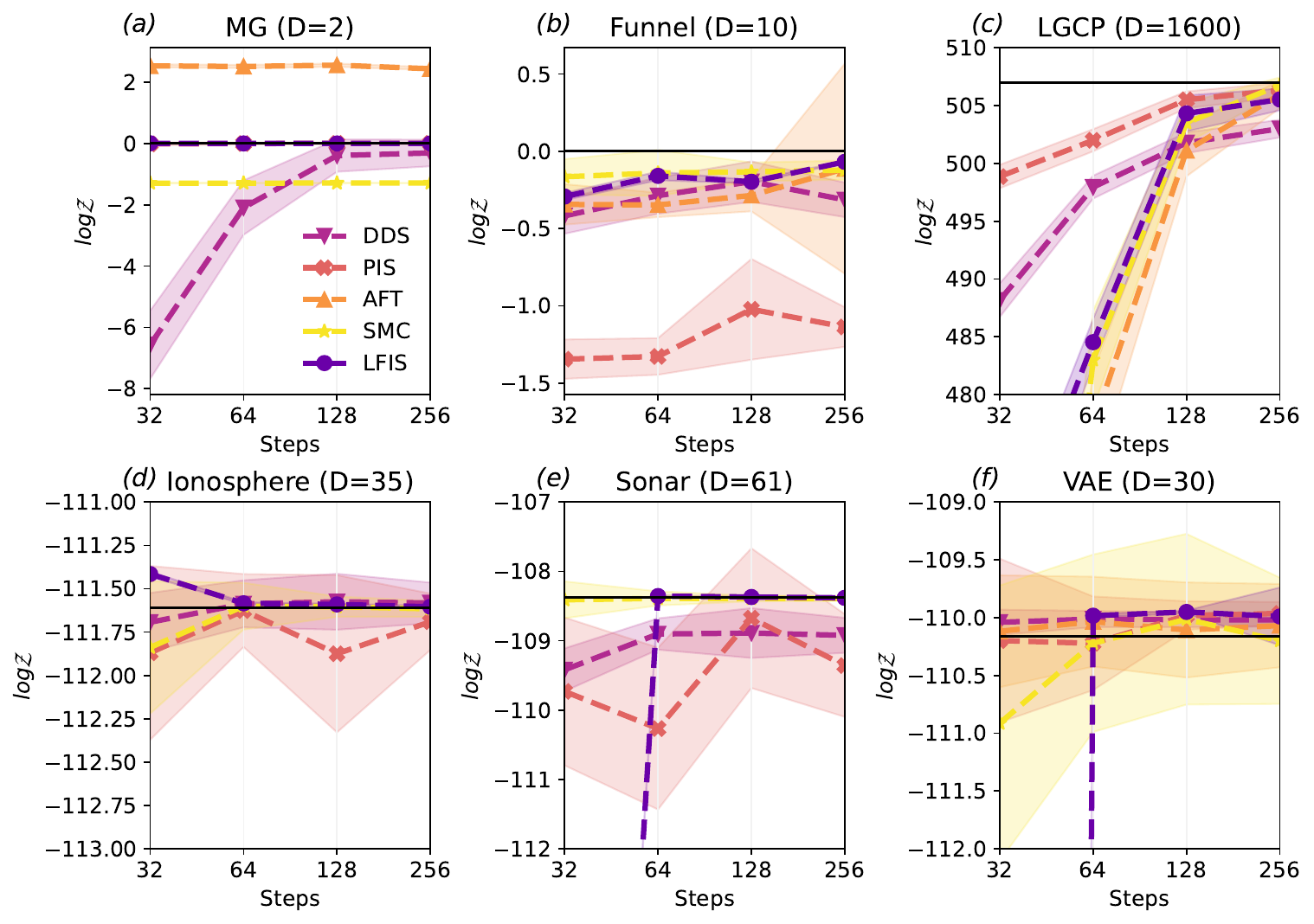}
    \caption{The effects of time steps on $\log\hat{\mathcal{Z}}$ esitmation}
    \label{fig:steps}
\end{figure}

Here, we provide the results of $T\in \left\{32,64,128,256\right\}$ for SMC (Table \ref{tab:SMC-all}), AFTMC (Table \ref{tab:AFTMC-all}), PIS (Table \ref{tab:PIS-all}), DDS (Table \ref{tab:DDS-all}), LFIS (Table \ref{tab:LFIS-all}), and a summary plot (Fig.~\ref{fig:steps}). To enforce uniformity, we ran each method 30 times independently and computed the summary statistics from the collected results. 

For the two transformation problems: Gaussian mixture and funnel distributions, we perform the numerical experiments under the assumption that the coverage of the distribution is unknown so that the prior/reference distribution is set to be the simplest uncorrelated normal distributions with zero mean and identity covariance matrix. For LFIS, this can be achieved by setting $\rho_\ast(0) = \mathcal{N}(0,I)$. For DDS, we set the $\sigma$ of the Ornstein–Uhlenbeck process to be 1, which will result in a reference distribution of $\mathcal{N}(0,I)$. For PIS, the uncontrolled reference distribution is determined by the end time of the stochastic process $T_\text{end}$ and the magnitude of the stochastic process \texttt{g\_coef}. Using the original settings in the PIS Github repo with \texttt{g\_coef}=$\sqrt{0.2}$, we set $T_\text{end}= 5.0$ to fix the uncontrolled reference distribution as $\mathcal{N}(0,I)$. By doing so, we impose a more challenging and practical set of problems for comparing different methods.

For the Bayesian problems, it is natural to choose the prior distribution as the initial distribution for LFIS. For DDS, to avoid tuning hyper-parameters, we directly use the results provided in the DDS Github repository \citep{DDSrepo}, which has the same numerical setup for computing the statistics as all other methods. For PIS, we can reproduce the statistics for most of the Bayesian problems by tuning the hyperparameters, except for the LGCP. For completeness, we use the statistics reported in the DDS GitHub repository \citep{DDSrepo} for table \ref{tab:PIS-all}.

\begin{table}[!t]
\begin{footnotesize}
\caption{All $\log\mathcal{Z}$ results of SMC with different tempering scale setting ($T$).}
\vskip 0.15in
\centering
\begin{tabular}{ccccccc}
\toprule
$T$  & MG ($D=2$) & Funnel ($D=10$) &  VAE ($D=30$) & Ionosphere ($D=35$) & Sonar ($D=61$) & LGCP ($D=1600$)\\ 
\midrule 
32 & $-1.29 \pm 0.015$ & $-0.17 \pm 0.11$ & $-110.91 \pm 1.19$ & $-111.84 \pm 0.38$ & $-108.41 \pm 0.27$ & $408.75 \pm 7.27$ \\
64 & $-1.29 \pm 0.013$ & $-0.14 \pm 0.15$ & $-110.22 \pm 0.77$ & $-111.60 \pm 0.14$ & $-108.39 \pm 0.10$ & $483.00 \pm 4.26$ \\
128 & $-1.29 \pm 0.010$ & $-0.13 \pm 0.061$ & $-110.01 \pm 0.74$ & $-111.60 \pm 0.059$ & $-108.39 \pm 0.04$ & $503.51 \pm 1.74$ \\
256 & $-1.29 \pm 0.006$ & $-0.12 \pm 0.062$ & $-110.20 \pm 0.55$ & $-111.62 \pm 0.046$ & $-108.39 \pm 0.035$ & $506.77 \pm 0.68$ \\
1024 & $-1.29 \pm 0.004$ & $-0.03 \pm 0.17$ & $-110.16 \pm 0.41$ & $-111.61 \pm 0.026$ & $-108.38 \pm 0.018$ & $506.96 \pm 0.24$ \\
\bottomrule
\end{tabular}
\label{tab:SMC-all}
\end{footnotesize}
\end{table}

\begin{table}[!t]
\begin{footnotesize}
\caption{All $\log\mathcal{Z}$ results of AFTMC with different tempering scale setting ($T$).}
\vskip 0.15in
\centering
\begin{tabular}{ccccccc}
\toprule
$T$  & MG ($D=2$) & Funnel ($D=10$) &  VAE ($D=30$) & Ionosphere ($D=35$) & Sonar ($D=61$) & LGCP ($D=1600$)\\ 
\midrule 
32 & $2.53 \pm 0.05$ & $-0.34 \pm 0.13$ & $-110.12 \pm 0.48$ & $-86.03 \pm 6.95$ & $-99.39 \pm 39.49$ & $406.90 \pm 7.96$ \\
64 & $2.51 \pm 0.05$ & $-0.34 \pm 0.079$ & $-110.03 \pm 0.39$ & $-118.46 \pm 64.28$ & $-97.82 \pm 65.72$ & $477.27 \pm 4.76$ \\
128 & $2.55 \pm 0.05$ & $-0.28 \pm 0.10$ & $-110.11 \pm 0.41$ & $-113.33 \pm 16.61$ & $-98.18 \pm 82.91$ & $501.11 \pm 2.23$ \\
256 & $2.44 \pm 0.05$ & $-0.11 \pm 0.68$ & $-110.07 \pm 0.36$ & $-121.63 \pm 16.37$ & $-104.80 \pm 68.31$ & $505.97 \pm 1.19$ \\
\bottomrule
\end{tabular}
\label{tab:AFTMC-all}
\end{footnotesize}
\end{table}

\begin{table}[!t]
\begin{footnotesize}
\caption{All $\log\mathcal{Z}$ results of PIS with different tempering scale setting setting ($T$). The statistics marked by $*$ are results from \citet{DDS} using the same numerical experimental setup.  }
\vskip 0.15in
\centering
\begin{tabular}{ccccccc}
\toprule
$T$  & MG ($D=2$) & Funnel ($D=10$) &  VAE ($D=30$) & Ionosphere ($D=35$) & Sonar ($D=61$) & LGCP ($D=1600$)\\ 
\midrule 
32 & $-0.016 \pm 0.045$ & $-1.35 \pm 0.13 $ & $-110.20 \pm 0.71$ & $-111.87 \pm 0.50$ & $-109.73 \pm 1.07$ & $498.86 \pm 1.02 *$ \\
64 & $-0.003 \pm 0.037$ & $-1.33 \pm 0.12$ & $-110.22 \pm 0.41$ & $-111.63 \pm 0.21$ & $-110.27 \pm 1.17$ & $502.00 \pm 0.96 *$ \\
128 & $0.002 \pm 0.030$ & $-1.02 \pm 0.33$ & $-109.98 \pm 0.12$ & $-111.88 \pm 0.45$ & $-108.67 \pm 1.01$ & $505.50 \pm 0.72 *$ \\
256 & $0.004 \pm 0.018$ & $-1.14 \pm 0.13$ & $-109.96 \pm 0.10$ & $-111.69 \pm 0.16$ & $-109.36 \pm 0.74$ & $506.34 \pm 0.63 *$ \\
\bottomrule
\end{tabular}
\label{tab:PIS-all}
\end{footnotesize}
\end{table}

\begin{table}[!t]
\begin{footnotesize}
\caption{All $\log\mathcal{Z}$ results of DDS with different tempering scale setting setting ($T$). The statistics marked by $*$ are results from \citet{DDS} using the same numerical experimental setup.}
\vskip 0.15in
\centering
\begin{tabular}{ccccccc}
\toprule
$T$  & MG ($D=2$) & Funnel ($D=10$) &  VAE ($D=30$) & Ionosphere ($D=35$) & Sonar ($D=61$) & LGCP ($D=1600$)\\ 
\midrule 
32 & $-6.57 \pm 1.11$ & $-0.42 \pm 0.12$ & $-110.04 \pm 0.11*$ & $-111.69 \pm 0.17 *$ & $-109.41 \pm 0.31*$ & $488.17 \pm 1.45*$ \\
64 & $-2.09 \pm 0.89$ & $-0.28 \pm 0.12$ & $-110.01 \pm 0.07*$ & $-111.59 \pm 0.14*$ & $-108.90 \pm 0.23*$ & $497.94 \pm 0.99*$ \\
128 & $-0.40 \pm 0.53$ & $-0.20 \pm 0.13$ & $-110.01 \pm 0.06*$ & $-111.58 \pm 0.16*$ & $-108.89 \pm 0.36*$ & $501.78 \pm 0.90*$ \\
256 & $-0.31 \pm 0.43$ & $-0.31 \pm 0.11$ & $-110.01 \pm 0.06*$ & $-111.58 \pm 0.12*$ & $-108.92 \pm 0.25*$ & $503.01 \pm 0.77*$ \\
\bottomrule
\end{tabular}
\label{tab:DDS-all}
\end{footnotesize}
\end{table}

\begin{table}[!t]
\begin{footnotesize}
\caption{All $\log\mathcal{Z}$ results of LFIS with different tempering scale setting setting ($T$).}
\vskip 0.15in
\begin{center}
\begin{tabular}{ccccc}
\toprule
&$T$  & MG ($D=2$) & Funnel ($D=10$) &  VAE ($D=30$) \\ 
\midrule 
&32 & $-0.064 \pm 0.065$ & $-0.43 \pm 0.050$ & $-220.70 \pm 2.23$ \\
{$\widehat{\log\mathcal{Z}}$} &64 & $-0.019 \pm 0.060$ & $-0.27 \pm 0.066$ & $-110.58 \pm 0.39$ \\
&128 & $-0.007 \pm 0.066$ & $-0.23 \pm 0.057$ & $-110.07 \pm 0.31$  \\
&256 & $0.004 \pm 0.059$ & $-0.11 \pm 0.075$ & $-109.95 \pm 0.25$  \\
\midrule 
& {32} &  {$0.006 \pm 0.005$} &  {$-0.29 \pm 0.023$} &  {$-215.64 \pm 0.87$} \\
{$\log\hat{\mathcal{Z}}$} & {64} &  {$0.002 \pm 0.003$} &  {$-0.16 \pm 0.028$} &  {$-109.98 \pm 0.01$} \\
& {128} &  {$-0.0001 \pm 0.004$} &  {$-0.20 \pm 0.011$} &  {$-109.95 \pm 0.01$}  \\
&  {256} &  {$-0.0002 \pm 0.004$} &  {$-0.07 \pm 0.003$} &  {$-109.99 \pm 0.08$}  \\
\midrule
&$T$  & Ionosphere ($D=35$) & Sonar ($D=61$) & LGCP ($D=1600$)\\ 
\midrule 
&32 & $-112.28 \pm 0.68$ & $-133.36 \pm 2.08$ & $470.13 \pm 5.57$ \\
{$\widehat{\log\mathcal{Z}}$} &64  & $-112.04 \pm 0.38$ & $-108.94 \pm 0.30$ & $489.31 \pm 3.20$ \\
&128 & $-111.67 \pm 0.26$ & $-108.60 \pm 0.31$ & $505.26 \pm 5.56$ \\
&256  & $-111.63 \pm 0.33$ & $-108.48 \pm 0.30$ & $505.43 \pm 3.00$ \\
\midrule 
& {32} &  {$-111.42 \pm 0.01$} & {$-130.45 \pm 0.233$} &  {$463.40 \pm 2.61$} \\
{$\log\hat{\mathcal{Z}}$} & {64} &  {$-111.58 \pm 0.006$} &  {$-108.36 \pm 0.011$} &  {$484.54 \pm 2.01$} \\
& {128}  &  {$-111.59 \pm 0.004$} &  {$-108.37 \pm 0.008$} &  {$504.32 \pm 1.56$} \\
&  {256}  &  {$-111.60 \pm 0.006$} &  {$-108.38 \pm 0.009$} &  {$505.53 \pm 0.95$} \\
\bottomrule
\end{tabular}
\label{tab:LFIS-all}
\end{center}
\end{footnotesize}
\end{table}

\subsection{Sequential Monte Carlo} \label{app:SMC}

Sequential Monte Carlo \citep{SMC} serves both as the reference method as well as the gold standard for those problems that are not analytically tractable. A series of similar studies including AFT \citep{AFT}, PIS \citep{zhang2021path}, and DDS \citep{DDS} took the same approach. We found that the SMC implementations in AFT and DDS were unnecessarily complicated. In particular, the leapfrog step size in the Hamiltonian Monte Carlo (HMC) kernel in those implementations depends on the progression of the tempered scales. Although progression-dependent step size was meticulously tuned, we found that a fixed step size with a sufficiently long HMC kernel and sufficiently high Effective Sample Size (ESS) threshold for triggering resampling can achieve similar results.

As such, we have streamlined the SMC implementation in this manuscript. The source codes, implemented in PyTorch, will be released if the manuscript is accepted for publication. Our SMC implementation is relatively simple: for each scale, the Monte Carlo step consists of $N_H$ HMC kernels, each of which has $L$ leapfrog steps, and each time step is fixed at $\delta$. Table \ref{tab:SMCspec} specifies these hyper-parameters of the SMC for each of the test problems.

\begin{table}[!h]
\begin{footnotesize}
\caption{Specification of our SMC implementations.} \label{tab:SMCspec}
\vskip 0.15in
\begin{center}
\begin{tabular}{cccccccc}
\toprule
  & MG & Funnel&  VAE  & Ionosphere  & Sonar  & LGCP  \\ 
\midrule 
$\delta$ & 0.02 & 0.02 & 0.2 & 0.02 & 0.02 & 0.2 \\
$L$  & 20 & 20 & 10 & 20 & 20 & 20 \\
$N_H$ & 10 & 10 & 2 & 10 & 10 & 2 \\
ESS threshold & 0.98 & 0.98 & 0.98 & 0.98 & 0.98 & 0.98 \\
\bottomrule
\end{tabular}
\end{center}
\end{footnotesize}
\end{table}

We also noticed a nuanced detail about applying SMC on LGCP. AIS and DDS both suggested performing SMC on a whitened representation of the LGCP problem, which was based on diagonalizing the covariance matrix of the problem by Cholesky decomposition. However, we found that for the tempering scale $T\gtrsim 256$, a whitened representation is not necessary. We provide a comparison of applying SMC on the whitened and un-whitened representations in Table \ref{tab:whitenOrNot}. As such, we only reported the SMC results on the unwhitened LGCP problem in the rest parts of our manuscript. 

\begin{table}[!h]
\begin{footnotesize}
\caption{SMC estimation of $\log \mathcal{Z}$ on whitened and unwhitened representation of the Log-Gaussian Cox Problem.} \label{tab:whitenOrNot}
\vskip 0.15in
\begin{center}
\begin{tabular}{ccc}
\toprule
 T & Whitened & Unwhitened  \\ 
\midrule 
32 & $ 506.90 \pm 0.12$ & $ 408.75 \pm 7.27$ \\
64 & $ 506.89 \pm 0.09$ & $ 483.00 \pm 4.26$ \\
128 & $ 506.87 \pm 0.068$ & $ 503.51 \pm 1.74$ \\
256 & $ 506.89 \pm 0.049$ & $ 506.77 \pm 0.68$ \\
1024 & $ 506.90 \pm 0.018$ & $ 506.96 \pm 0.24$ \\
\bottomrule
\end{tabular}
\end{center}
\end{footnotesize}
\end{table}

\section{Ablation study of Annealed Flow Transport Monte Carlo Sampler } \label{app:AFT-noMC}

We tested the AFTMC sampler (with 256 scales) without the MC kernel on the Gaussian Mixture, Funnel, Variational Auto-Encoder, and Log-Gaussian Cox Process problems. We juxtapose the results of (1) SMC with $T=1024$ as the gold standard, (2) SMC with $T=256$ as a reference, (3) AFT with a Monte Carlo kernel and $T=256$, and (4) AFT without the Monte Carlo kernel and $T=256$ in Table \ref{tab:AFT-noMC}. The results indicate that the MC kernel, instead of the flow-transport kernel, is the major functioning component in estimating the marginal likelihood $\log \mathcal{Z}$. The AFTMC algorithm with only the MC kernel is functionally identical to SMC.  The numerical experiments on these datasets suggest that the flow transport, implemented as a normalizing flow and parametrized by variational inference, does not function independently from the MCMC steps, which still need to be designed and fine-tuned akin to typical SMCs. This contrasts with the proposed LFIS, which solely depends on the flow for transporting samples without designing additional MCMC kernels.

\begin{table}[!t]
\caption{Ablation study on the effect of Monte Carlo kernel in Annealed Flow Transport Monte Carlo sampler \citep{AFT} estimating $\log \mathcal{Z}$.}
\vskip 0.15in
\begin{small}
\begin{center}
\begin{tabular}{lcccc}
\toprule
Model  & MG ($D=2$) & Funnel ($D=10$) & VAE ($D=30$) & LGCP ($D=1600$)\\ 
\midrule 
SMC (1024) & $-1.29 \pm 0.0046$ & $-0.034\pm 0.17$& $ -110.16 \pm 0.41$ &  $506.96 \pm 0.24$\\
SMC (256) & $-1.29 \pm 0.0062$ & $-0.12\pm 0.062$& $-110.20 \pm 0.54$ &  $506.77 \pm 0.68$ \\
AFT with MC (256) & $2.44 \pm 0.05$ & $-0.11 \pm 0.68$ & $-110.07 \pm 0.36$ & $505.97 \pm 1.19$\\
AFT without MC (256) & $2.94 \pm 0.39$ & $-0.79 \pm 0.40$ & $-288.18 \pm 58.28 $& $-826.73 \pm 8.96$\\
\bottomrule
\end{tabular}
\label{tab:AFT-noMC}
\end{center}
\end{small}
\end{table}

\begin{table}[!t]
\centering
\begin{small}
\caption{Adjustable meta-parameters and objects in a range of methods. Markers $\checkmark$ indicate the adjustable objects that the user must specify. We remark that within the MC kernel there could be multiple adjustable parameters, e.g., see Sec.~\ref{app:SMC} and Table \ref{tab:SMCspec}.}
\vskip 0.15in
\begin{tabular}{lcccccc}
\toprule
Adjustable parameters/objects  & SMC & AFTMC & CR-AFTMC & PIS & DDS & LFIS \\ 
\midrule 
Number of time steps/scales $T$ & $\checkmark$ & $\checkmark$& $\checkmark$& $\checkmark$& $\checkmark$& $\checkmark$ \\
Terminal time &  & & & $\checkmark$ &  &  \\
Neural network (per time step/scale)  &  & $\checkmark$& $\checkmark$& $\checkmark$& $\checkmark$& $\checkmark$ \\
Monte Carlo kernel (per time step/scale) & $\checkmark$ & $\checkmark$& $\checkmark$& && \\
Reference process $\tilde{\rho}_\ast(x,t)$ & & & & $\checkmark$  & $\checkmark$  & \\
Annealed path of distributions $\tilde{\rho}_\ast(x,t)$ & $\checkmark$ & $\checkmark$& $\checkmark$& & & $\checkmark$ \\
Resampling threshold & $\checkmark$ & $\checkmark$& $\checkmark$& & &  \\
Initial distribution & $\checkmark$& $\checkmark$& $\checkmark$& & & $\checkmark$\\
\bottomrule
\end{tabular}
\label{tab:modelComplexity}
\end{small}
\end{table}

\end{document}